%% file: arxiv_bytedance.tex
\documentclass[]{bytedance}
\usepackage[toc,page,header]{appendix}

\usepackage{minitoc}
\usepackage{amsfonts}
\usepackage{amssymb}
\usepackage{tabularx}
\usepackage{listings}
\usepackage{xcolor}
\usepackage{cancel}

\usepackage{tabulary,multirow,xspace}
\usepackage{fixmath,mathtools,nicefrac,mmstyle}
\usepackage{subcaption}
\usepackage{caption}
\usepackage{float}
\usepackage{wrapfig} 
\usepackage[misc]{ifsym} 
\usepackage{colortbl}

\usepackage{wrapfig}
\usepackage{multicol}
\usepackage[most]{tcolorbox}
\usepackage{pifont}

\definecolor{codegreen}{rgb}{0,0.6,0}
\definecolor{codegray}{rgb}{0.5,0.5,0.5}
\definecolor{codepurple}{rgb}{0.58,0,0.82}
\definecolor{backcolour}{rgb}{0.95,0.95,0.92}
\definecolor{boxblue}{RGB}{57,89,163}
\definecolor{boxbluebg}{RGB}{230,237,250} 

\lstdefinestyle{mystyle}{
    backgroundcolor=\color{backcolour},   
    commentstyle=\color{codegreen},
    keywordstyle=\color{magenta},
    numberstyle=\tiny\color{codegray},
    stringstyle=\color{codepurple},
    basicstyle=\ttfamily\footnotesize,
    breakatwhitespace=false,         
    breaklines=true,                 
    captionpos=b,                    
    keepspaces=true,                 
    numbers=none,                    
    numbersep=5pt,                  
    showspaces=false,                
    showstringspaces=false,
    showtabs=false,                  
    tabsize=2
}
\definecolor{mygray1}{gray}{.95}
\definecolor{mygray2}{gray}{.9}
\definecolor{mygray3}{gray}{.95}
\usepackage{pifont}
\newcommand{\cmark}{\ding{51}}%
\newcommand{\xmark}{\ding{55}}%

\newlength\savewidth
\newcolumntype{x}[1]{>{\centering\arraybackslash}p{#1pt}}

\newcommand{\app}{\raise.17ex\hbox{$\scriptstyle\sim$}}

\definecolor{merit_red}{RGB}{255,46,99}
\definecolor{merit_dark}{RGB}{37,42,52}
\definecolor{merit_blue}{RGB}{8,217,214}
\definecolor{merit_gray}{RGB}{156,156,156}

\title{DenseWorld-1M: Towards Detailed Dense Grounded Caption in the Real World}

{
\small 
\author[\dagger\star]{Xiangtai Li$^{1}$}
\author[\star]{Tao Zhang$^{1}$}
\author[\star]{Yanwei Li$^{1}$}
\author[]{Haobo Yuan$^{2}$}
\author[]{Shihao Chen$^{2}$}
\author[]{Yikang Zhou$^{2}$}
\author[]{Jiahao Meng$^{3}$}
\author[]{Yueyi Sun$^{3}$}
\author[]{Shilin Xu$^{3}$}
\author[]{Lu Qi$^{1}$}
\author[]{Tianheng Cheng$^{1}$}
\author[]{Yi Lin$^{1}$}
\author[]{Zilong Huang$^{1}$}
\author[]{Wenhao Huang$^{1}$}
\author[]{Jiashi Feng$^{1}$}
\author[]{Guang Shi${^1}$}
}
\affiliation[]{ {ByteDance Seed$^{1}$} \quad  Wuhan University$^{2}$ \quad Peking University$^{3}$ }
\contribution[\star]{Equal technical contributions to the work.}

\abstract{
Multimodal Large Language Models (MLLMs) demonstrate a complex understanding of scenes, benefiting from large-scale and high-quality datasets.
Most existing caption datasets lack the ground locations and relations for visual entities.
Several grounded caption datasets face the problems of missing detailed descriptions, relations, and massive object descriptions on high-resolution images.
To fill this gap for the community, we present DenseWorld-1M, the first massive, detailed, dense grounded caption dataset in the real world. 
We design a three-stage labeling pipeline, containing open-world perception, detailed object caption generation, and dense caption merging.
The first stage obtains entity-level masks and labels. 
The second stage generates the object-level, detailed captions with the guidance of masks and labels from the first stage.
The final stage merges object captions and masks into spatial and relational dense captions.
To accelerate the labeling process and improve caption quality, we present two VLM models: the Detailed Region Caption model and the Spatial Caption Merging model.
Extensive experiments on various settings, including vision-language understanding, visual grounding, and region caption generation, demonstrate the effectiveness of our DenseWorld-1M dataset and labeling models.
The datasets and models will be released at \url{https://github.com/lxtGH/DenseWorld-1M}.
}

\date{\today}
\correspondence{Xiangtai Li: \email{xiangtai.li@bytedance.com}}

\begin{document}

\maketitle


\input{sec/1_intro}
\input{sec/2_related_work}
\input{sec/3_method}

\input{sec/4_exp}
\input{sec/5_conclusion}



\input{sec/X_suppl}

\clearpage
\bibliographystyle{plainnat}
\bibliography{main}



\end{document}

%% file: sec/1_intro.tex
\section{Introduction}
\label{sec:intro}

Current state-of-the-art MLLMs~\cite{zhu2025internvl3,chen2024internvl2_5,bai2025qwen2,wang2024qwen2,lin2023video, long-vita, vita, videorag, mllm-selector, wu2024controlmllm, luo2024feast, mmict, luo2023cheap, feng2024align, li2023rain, han2024free, bi2024forest, rang2025eve, zhang2025enhancing, zhang2024seeing, seed2025seed1_5vl} benefit from massive, diverse, high-quality training datasets for different stages, including pre-training, supervised fine-tuning, and post-training.
Several research works~\cite{tong2024cambrian,chen2024internvl2_5,zhu2025internvl3} indicate the essential roles of datasets in achieving the state-of-the-art performance on various MLLM benchmarks.
Recently, there have been more urgent requirements~\cite{rasheed2024glamm,zhang2024omg} for a fine-grained understanding of MLLMs in the real world, as this is a vital step in enabling machines to interpret and interact with diverse visual information, just like humans.
However, existing MLLM datasets~\cite{fu2023mme,chen2015microsoft,ai2d,yuan2025sa2va} lack detailed, dense object captions, grounded captions, and object location information.
For example, several datasets~\cite{li2024densefusion, yuan2024osprey, cui2025comprehensive, chen2024allava, lin2024draw, guo2024mammoth} only contain detailed captions without masks or relations between entities. 
In particular, DenseFusion-1M~\cite{li2024densefusion} only contains dense text without spatially grounded locations (masks, boxes), making it hard to carry out region-level understanding.

On the other hand, several region-level datasets always lack detailed caption annotations. 
In particular, both referring datasets and region caption datasets~\cite{rasheed2024glamm} have issues, including short descriptions, missing background context annotations, and incomplete fine-grained object descriptions. 
%
In particular, several datasets present dense grounded caption annotations. However, these datasets still have shortcomings, including low-resolution images, simple scenes, missing detailed object descriptions, and missing spatial relations.
%
(See the Appendix~\ref{app_sec:dataset_details} for the specific examples in previous datasets.)
In this work, we aim to present a new dataset that addresses the aforementioned issues.
The Tab.~\ref{tab:dataset_comparison} summarizes the differences between our dataset (DenseWorld-1M) and the previous datasets on various aspects.
To the best of our knowledge, this is the first large-scale, detailed, dense, grounded caption dataset in the common scene.

\begin{figure*}[t]
    \centering
    \includegraphics[width=0.78\linewidth]{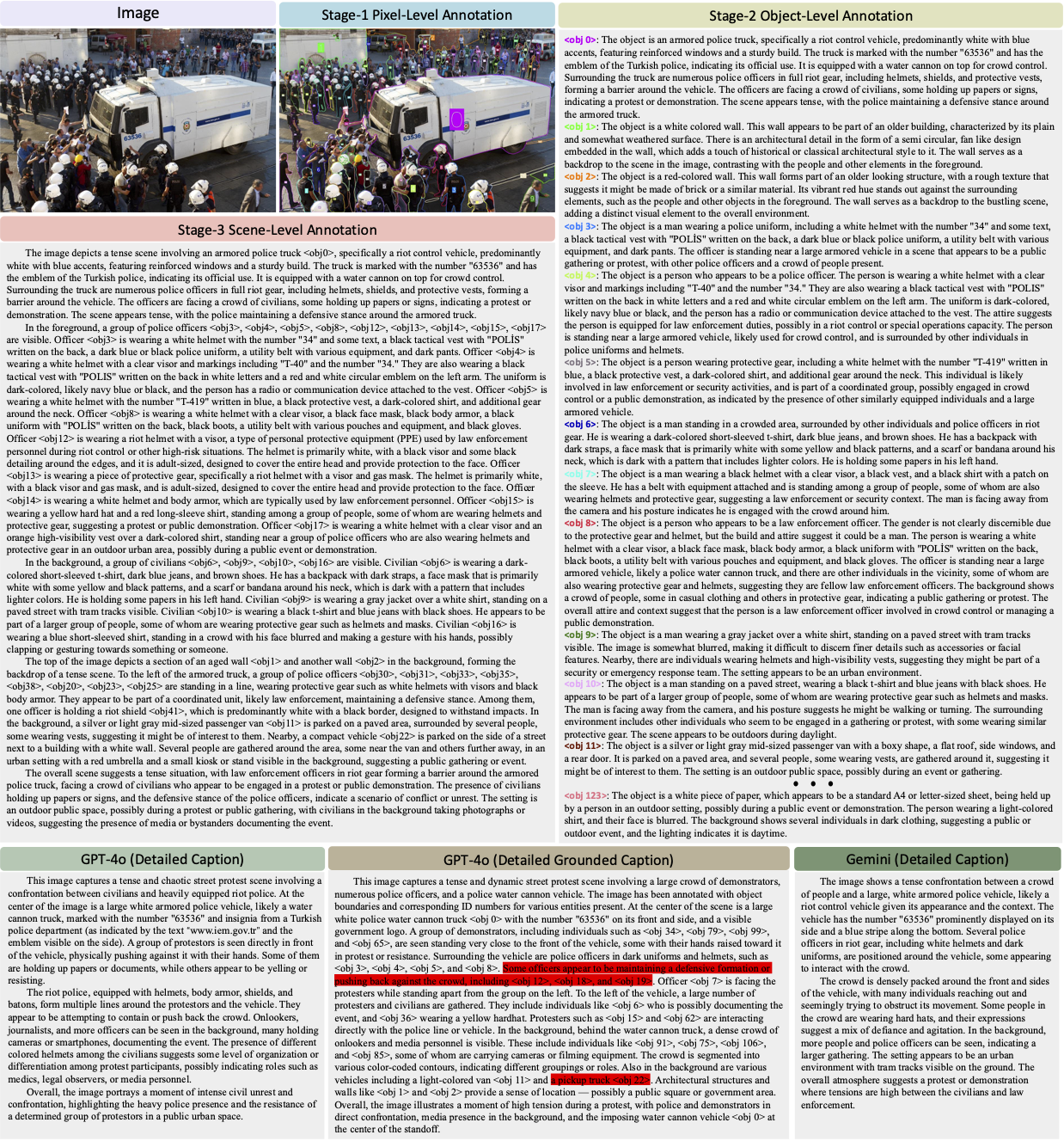}
    \caption{\small \textbf{DenseWorld-1M Annotation Example.} DenseWorld-1M contains extremely dense and detailed grounded captions, with three-stage outputs. Current private models, such as GPT-4 and Gemini, are unable to generate such captions, even when provided with pixel-level tags as visual prompts. The inconsistencies between the text and object ID numbers in the detailed grounded captions generated by GPT-4o are highlighted in red. Best view it in color.} 
    \label{fig:teaser}
\end{figure*}

These rich semantic annotations need huge and expensive human annotations.
Thus, we aim to leverage existing state-of-the-art models to develop an automatic pipeline to scale the data labeling process.
Motivated by recent vision foundation models and open-sourced MLLMs~\cite{bai2025qwen2,shen2024aligning,kirillov2023segment}, we use the specialized visual segmentation model outputs (masks) as an intermediate bridge to generate detailed object captions.
Our pipeline can be divided into three stages: pixel grouping, detailed object caption generation, and dense ground caption merging. 
In the first stage, we use visual foundation models to generate dense entity masks, where we design a mask merging and refinement pipeline to improve mask quality.
At the second stage, with the generated mask and visual prompts from the first stage, we leverage the state-of-the-art MLLMs to develop the detailed object captions.
In particular, we design a cropping and fusing pipeline, along with a verification model, to refine object captions.
In the third stage, we present a caption merging pipeline that merges the detailed captions from the earlier stage with spatial awareness.
%

To accelerate the labeling pipeline in stage-2 and stage-3, where the computation costs are more significant with multiple MLLM inferences, we present two tuned models: a detailed region caption model (DRC) and a spatial caption merging model (SPM). 
For the former, we present a new token injection design to fuse two-stream tokens.
For the latter, with object tags and stage-3 ground truth, we fine-tune one MLLM to merge the detailed object captions.
Both models avoid repeated computation in the previous stage-2 and stage-3, as shown in Sec.~\ref{sec:exp}.
Both models are in the labeling loop, where 40\% data are obtained by the two models.
%
We conduct extensive experiments to verify the effectiveness of our dataset and two models.
In particular, we find that even state-of-the-art MLLMs, including both image-level variants and pixel-level variants, can still be improved. 
In particular, for grounded caption generation and region caption tasks, we achieve over 3
As shown in Fig.~\ref{fig:teaser}, compared with private MLLMs (GPT-4o and Gemini-Flash-2.0), our dataset provides a more detailed understanding of complex scenes.

In summary, our contributions are listed as follows:
\begin{itemize}
\item We present the first large-scale, detailed, dense grounded captions, DenseWorld-1M, for the MLLM community. 
\item We develop a bottom-up, three-stage dataset generation pipeline that leverages existing state-of-the-art perception models and MLLMs to automatically generate dense, grounded captions, masks, and labels.
\item We propose two fine-tuned MLLMs: DRC and SCM, to accelerate the labeling process.
\item Extensive experiments show the effectiveness of DenseWorld-1M on over 10 different datasets. Comparison results also show the effectiveness of DRC and SCM.
\end{itemize}

\begin{table}[t]
    \centering
    \caption{\small \textbf{Summary and Comparison with Existing Dense Captions and Grounding Datasets.} We compare existing works from aspects including: $^1$DC: Dense Captions, $^2$GC: Grounded Caption, $^3$OC: Object Caption,  $^4$DOC: Detailed Object Captions (over 150 text tokens per obejcts), $^5$DGC: Dense Grounded Caption, $^6$SR: Spatial Relation, $^7$HRI: High Resolution Images (over 4K resolution images), $^8$LV: Large Vocabulary (classes over 20k), and $^9$PA: Pixel-level Annotations. }
    \label{tab:dataset_comparison}
    \resizebox{\textwidth}{!}{
    \begin{tabular}{r|r|ccccccccc|r}
    \toprule
    \multirow{1}{*}{\textbf{Dataset}} &  {\textbf{Year}} & {\textbf{DC}} &  
  \textbf{GC} & \textbf{OC} & \textbf{DOC} & \textbf{DGC} & \textbf{SR} & \textbf{HRI} & \textbf{LV} & \textbf{PA} & {\textbf{Images}}
    \\
    \midrule
    GLaMM-GCG~\cite{rasheed2024glamm} & \textcolor{gray}{{\small 2023}}  & \textcolor{merit_blue}{\xmark}  & \textcolor{merit_red}{\cmark} & \textcolor{merit_blue}{\xmark} & \textcolor{merit_blue}{\xmark} & \textcolor{merit_blue}{\xmark} & \textcolor{merit_blue}{\xmark} & \textcolor{merit_red}{\cmark} & \textcolor{merit_red}{\cmark} &  \textcolor{merit_red}{\cmark} & 214K \\
    ShareGPT4V~\cite{chen2024sharegpt4v} & \textcolor{gray}{{\small 2023}}  &  \textcolor{merit_red}{\cmark} & \textcolor{merit_red}{\cmark} & \textcolor{merit_blue}{\xmark} & \textcolor{merit_blue}{\xmark} & \textcolor{merit_blue}{\xmark} & \textcolor{merit_blue}{\xmark} & \textcolor{merit_blue}{\xmark} & \textcolor{merit_blue}{\xmark} & \textcolor{merit_blue}{\xmark} &  1M \\
    Osprey~\cite{yuan2024osprey}  & \textcolor{gray}{{\small 2023}} & \textcolor{merit_blue}{\xmark} & \textcolor{merit_blue}{\xmark} &  \textcolor{merit_red}{\cmark} & \textcolor{merit_blue}{\xmark} & \textcolor{merit_blue}{\xmark} & \textcolor{merit_blue}{\xmark} & \textcolor{merit_blue}{\xmark} &\textcolor{merit_blue}{\xmark} & \textcolor{merit_red}{\cmark} & <700K
    \\
    MUSE~\cite{ren2024pixellm}  & \textcolor{gray}{{\small 2024}} & \textcolor{merit_blue}{\xmark} & \textcolor{merit_blue}{\xmark} & \textcolor{merit_red}{\cmark} & \textcolor{merit_blue}{\xmark} & \textcolor{merit_blue}{\xmark} & \textcolor{merit_blue}{\xmark} & \textcolor{merit_blue}{\xmark} & \textcolor{merit_blue}{\xmark} & \textcolor{merit_red}{\cmark} &246K 
    \\
    DenseFusion~\cite{li2024densefusion}  & \textcolor{gray}{{\small 2024}} & \textcolor{merit_red}{\cmark} & \textcolor{merit_blue}{\xmark} & \textcolor{merit_blue}{\xmark} & \textcolor{merit_blue}{\xmark} & \textcolor{merit_blue}{\xmark} & \textcolor{merit_blue}{\xmark} & \textcolor{merit_red}{\cmark} & \textcolor{merit_red}{\cmark} & \textcolor{merit_blue}{\xmark} &1M \\
    COCONut-PanCap~\cite{deng2025coconut}  & \textcolor{gray}{{\small 2024}} & \textcolor{merit_red}{\cmark} & \textcolor{merit_red}{\cmark} & \textcolor{merit_red}{\cmark} & \textcolor{merit_blue}{\xmark} & \textcolor{merit_blue}{\xmark} & \textcolor{merit_blue}{\xmark} & \textcolor{merit_blue}{\xmark} & \textcolor{merit_blue}{\xmark} & \textcolor{merit_red}{\cmark} &118K \\
    Pix2Cap-COCO~\cite{you2025pix2cap} & \textcolor{gray}{{\small 2024}} & \textcolor{merit_red}{\cmark} & \textcolor{merit_red}{\cmark} & \textcolor{merit_blue}{\xmark} & \textcolor{merit_blue}{\xmark} & \textcolor{merit_blue}{\xmark} & \textcolor{merit_blue}{\xmark} & \textcolor{merit_blue}{\xmark} & \textcolor{merit_blue}{\xmark} & \textcolor{merit_red}{\cmark} &118K \\
    \midrule
    DenseWorld-1M (Ours) & \textcolor{gray}{{\small 2025}} & \textcolor{merit_red}{\cmark} & \textcolor{merit_red}{\cmark} & \textcolor{merit_red}{\cmark} & \textcolor{merit_red}{\cmark} & \textcolor{merit_red}{\cmark} & \textcolor{merit_red}{\cmark} & \textcolor{merit_red}{\cmark}  & \textcolor{merit_red}{\cmark} & \textcolor{merit_red}{\cmark} & 1M
    \\
    \bottomrule              
\end{tabular}
}
\end{table}

%% file: sec/2_related_work.tex
\section{Related Work}
\label{sec:related_work}


\noindent
\textbf{Multi-modal Datasets.} Earlier works~\cite{schuhmann2022laion,sharma2018conceptual_cc3m,changpinyo2021conceptual_CC12m,krishna2017visual,thomee2016yfcc100m} mainly explore image-text pairs, such as LAION~\cite{schuhmann2022laion}, CC12M~\cite{changpinyo2021conceptual_CC12m}, Visual Genome~\cite{krishna2017visual}, etc. 
These datasets facilitate the development of vision-language pre-training~\cite{radford2021learning} and image/video captioning~\cite{chen2015microsoft,krishna2017dense}.
Meanwhile, several datasets and benchmarks~\cite{fu2023mme, mmbench, docvqa, mluv, zhu2023genimage, fei2025path} focus on visual question answering tasks.
With the rapid progress on LLMs and MLLMs, the scale and diversity of multi-modal datasets also increase. 
Several large-scale caption datasets, including ShareGPT-4V~\cite{chen2024sharegpt4v}, ImageInWoods~\cite{garg2024imageinwords}, DenseFusion-1M~\cite{li2024densefusion}, and DOCCI~\cite{onoe2404docci}, are also proposed, containing detailed and dense captions.
Most are used for pre-training and the joint SFT process for MLLMs.
Meanwhile, several works explore the grounded captions and region-level caption datasets, such as GLaMM~\cite{rasheed2024glamm} and all-seeing datasets~\cite{wang2023all,wang2024all}. 
These datasets are used for region-level MLLM tasks~\cite{refcocog,lai2024lisa}, including visual grounding and region captions.
Our work, DenseWorld-1M, presents a large-scale, high-resolution, detailed grounded caption dataset.
We develop a three-stage annotation pipeline that eliminates the need for human annotations and employs a model-in-the-loop design to accelerate the labeling process.
Extensive experiments show the effectiveness of our dataset over various benchmarks and baselines.

\noindent
\textbf{Multi-modal Large Language Models.} Recent mainstream state-of-the-art MLLMs~\cite{bai2025qwen2,chen2024expanding,liu2024llavanext,wang2024world,wang2025vgr} adopt a connected architecture, including one vision encoder, one project module, and one LLM. 
Benefit from large-scale, high-quality, diverse datasets training, including pre-training, SFT, and post-training, these models achieve stronger performance on multiple benchmarks. 
In addition, they also adopt various designs~\cite{liu2024llavanext,tong2024cambrian,zhang2024llavauhdv2,li2024mini_gemini} to handle the high-resolution images for detailed image understanding.
Meanwhile, several works~\cite{lai2024lisa,yuan2025sa2va,rasheed2024glamm,zhang2024omg,zhang2025pixel,yan2024visa} also study region-level grounding, referring segmentation, and grounded caption generation, with MLLM architecture.
However, these models still cannot generate detailed captions with grounded masks. 
These features are essential for the MLLMs to understand the real world, since the model can generate highly detailed captions with each precise entity in the scene, making the interaction easier.
Our work fills this gap by providing the community with a new dataset and pipelines.

\noindent
\textbf{Pixel-level Understanding.} Previous works~\cite{chen2017deeplab,he2017mask,girshick2015fast,li2020semantic,carion2020end,li2024transformer} explore specific models for segmentation and object detection. 
More recently, unified transformer-based models~\cite{yuan2022polyphonicformer,li2024omg,li2023tube,wang2023images,liu2024grounding,li2023panopticpartformerpp,chen2023generalist,xu2025rapsam}, have emerged to handle multiple dense prediction tasks within a single framework, demonstrating strong generalization across modalities. 
In contrast, our approach enjoys the progress of open-vocabulary vision models~\cite{zhang2024recognize,yuan2024open,shen2024aligning,ding2023open,wu2023clipself} to construct a flexible mask prediction pipeline that is not restricted to a single model type or task. 
Importantly, our method is complementary to existing perception models, as our primary objective is to generate rich, dense captions for multi-modal large language models (MLLMs), extending the role of pixel-level understanding beyond traditional perception.

%% file: sec/3_method.tex

\section{DenseWorld-1M Dataset}
\label{sec:denseworld_dataset}

\noindent
\textbf{Motivation.} Our goal is to build a detailed, dense, grounded caption dataset that can benefit a lot of downstream tasks such as VLM pretraining, text-driven grounding, and even O3-like agents that automatically check details. 
However, this is extremely challenging, and it's impossible to achieve this goal directly through zero-shot prompting of SOTA open-sourced MLLMs~\cite{zhu2025internvl3, li2024densefusion, bai2025qwen2} or closed-source SOTA models like GPT-4o~\cite{hurst2024gpt4o} and Gemini~\cite{team2023gemini} (See Fig.~\ref{fig:teaser} and Fig.~\ref{fig:caption_demos}).
\textit{First}, these models tend to focus only on the main objects in a scene rather than exhaustively describing all information in the entire image, as captions in current caption datasets do. \textit{Second}, these models struggle to accurately align text with corresponding object masks or bounding boxes in the final image caption.

\noindent
\textbf{Overall Pipeline.} To achieve this goal, we break down the task into three stages: pixel-level labeling (stage-1), object-level labeling (stage-2), and scene-level labeling (stage-3). 
In the first stage, we generate precise entity-level segmentation masks~\cite{qi2022open,kirillov2019panoptic} for each image to decompose a complex scene into multiple objects, where both foreground and background object masks are obtained.
In the second stage, we prompt SOTA MLLMs to focus only on a single object, thereby generating accurate and as exhaustive as possible descriptions for each object. 
In the third stage, we constrain the model through both visual prompts (highlighted edges and object IDs displayed on the image) and text prompts (detailed descriptions of each object generated in stage-2) to jointly produce a detailed grounded image caption with high consistency between text and grounding masks.

\noindent
\textbf{DenseWorld-1M Dataset Meta-Info.} Our original images have three sources, including SAM-1B~\cite{kirillov2023segment} (high resolution images), Object-365~\cite{shao2019objects365} (common object in the scene), and V3Det~\cite{wang2023v3det} (dataset with large vocabulary classes). Due to the limited pages, we put the meta information of DenseWorld-1M in the Appendix~\ref{app_sec:dataset_details}.

\subsection{Stage-1: Pixel-level Labeling}
\label{sec:method_stage_1}

\begin{figure}[t]
\hsize=\textwidth
\centering
\includegraphics[width=1.0\textwidth]{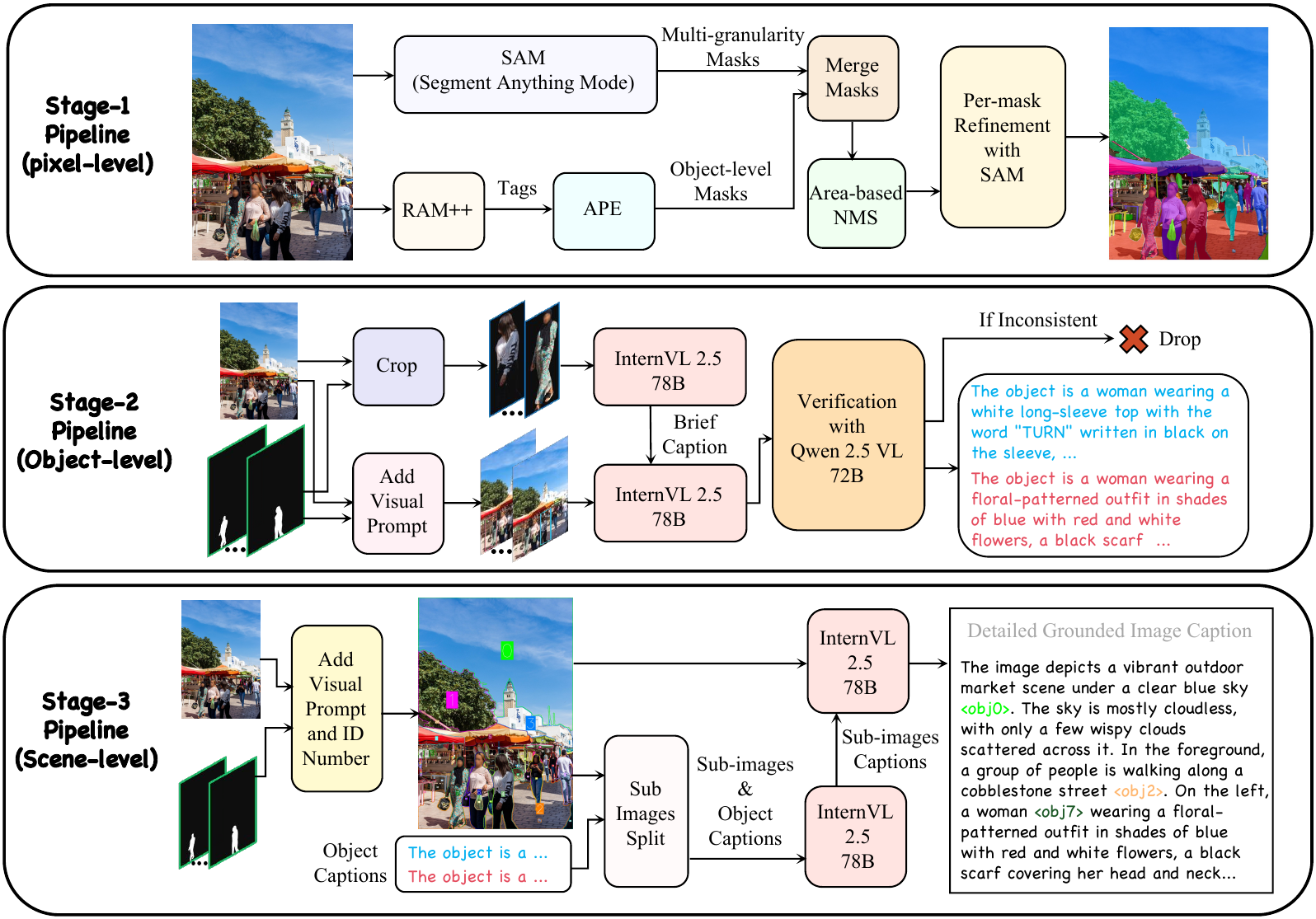}
\caption{\small{\textbf{DenseWorld-1M labeling pipeline}. We present a three-stage pipeline, including stage-1 for pixel-level mask generation, stage-2 for object-level detailed caption, and stage-3 for scene-level detailed dense grounded caption. Note that there are no human costs in the loop.}}
\label{fig:pipeline}
\end{figure}

As shown at the top of Fig.~\ref{fig:pipeline}, we integrate visual foundation models SAM~\cite{kirillov2023segment} and APE~\cite{shen2024aligning} and carefully design a filtering and refinement post-processing workflow to generate high-quality segmentation results. 
First, we use RAM++~\cite{zhang2024recognize} to generate object tags, which serve as an open vocabulary list to help APE generate panoptic segmentation masks. 
Meanwhile, we use SAM's segment anything mode to generate multi-granularity masks to supplement potential omissions produced by APE.
The union of segmentation results from APE and SAM provides excellent recall. 
However, it contains many duplicates, such as people and their upper bodies. 
To handle this, we first performed a merge process on the masks collection. 
If one mask falls completely within another mask and the IOU between these two masks is greater than 0.5, we merge these two masks and only keep the one with the largest area. 
Then, NMS is applied to filter duplicate masks, and since there is a refinement process afterward, we prioritize areas from large to small rather than confidence scores. 
After the merge and NMS process, there is no overlap between object masks, but there are still masks with low segmentation quality, such as fragmentation. 
Therefore, we refine each mask individually through SAM. Specifically, we use the current mask to regenerate multiple point prompts to prompt SAM to refine the mask, and select the one with the highest confidence score as the final output.
As shown in Fig.~\ref{fig:pipeline}, the stage-1 annotation pipeline can decompose complex scenes into different objects, while not containing numerous meaningless, trivial objects like those in the SAM dataset~\cite{kirillov2023segment}.

\subsection{Stage-2: Object-level Labeling}
\label{method:stage_2}

As shown in the middle of Fig.~\ref{fig:pipeline}, we prompt the SOTA MLLM~\cite{chen2024internvl2_5} to generate detailed descriptions for each object segmented in stage-1. 
We find that both general models~\cite{chen2024internvl2_5, bai2025qwen2} and specially designed region caption models~\cite{yuan2024osprey, rasheed2024glamm} struggle to accurately identify the referenced objects (often being influenced by main scene objects and similar objects) in complex scenes (especially SAM images) through visual prompts alone. 
To address this challenge, we first crop out the object using its mask. 
The cropped image of the individual object is then sent to InternVL-2.5 78B~\cite{chen2024internvl2_5} to generate a brief object description, focusing mainly on the object's category and primary appearance features. 
%
However, cropped images containing only a single object lose a lot of information, such as relationships with surrounding objects and ambiguous information that requires reasoning based on the surrounding context. 
Therefore, we generate detailed object descriptions by prompting InternVL-2.5 78B with both the visual prompt overlaid on the original image and the generated brief object caption as a text prompt. 
These descriptions include not only the detailed object appearance but also relationships with surrounding objects. 
Finally, we adopt Qwen2.5-VL 72B~\cite{bai2025qwen2} as a verification model to filter out all inconsistent object captions, ensuring high accuracy of the retained object captions.

\begin{figure}[t]
\hsize=\textwidth
\centering
\includegraphics[width=1.00\textwidth]{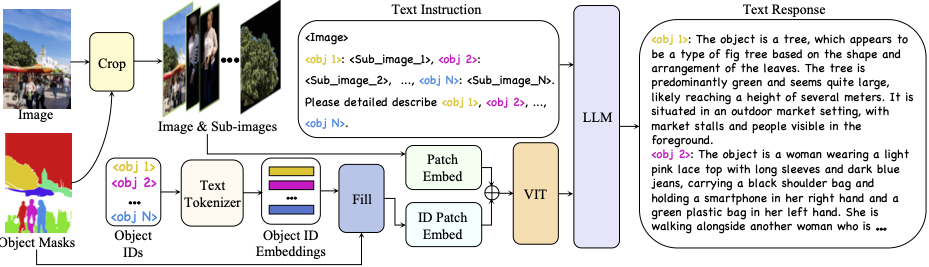}
\caption{\small{\textbf{The proposed Detailed Region Caption model.}} DRC combines both visual patch embedding and ID patch embedding to generate a more fine-grained and accurate description of object captions.}
\label{fig:object_cap_model}
\end{figure}

\noindent
\textbf{Using DRC.} Since the stage-2 pipeline calls large-sized models multiple times to generate object descriptions to ensure accuracy, this brings enormous computational costs. 
Therefore, considering the low-cost continuous scaling of datasets, we design and train a small 3B model using the 600k images data generated by the above pipeline. 
The details of the 3B detailed region caption model (DRC) are introduced in Sec.~\ref{sec:detailed_region_caption_model}. The DRC is much faster than InternVL-2.5 78B and can label multiple objects simultaneously, thus providing a 3-fold acceleration.


\subsection{Stage-3: Scene-level Labeling}

For simple scenes (with fewer than 15 objects), using MLLM directly based on object captions generated in stage-2 and images with visual prompts can effectively produce detailed grounded image captions. 
However, for complex scenes (with more than 15 objects), MLLM tends to retain only a few main objects. 
To address this challenge, as shown at the bottom of Fig.~\ref{fig:pipeline}, we divide complex scene images into multiple sub-images to reduce scene complexity. 
We use InternVL-2.5 78B~\cite{chen2024internvl2_5} to generate detailed grounded captions for each sub-image, and then use these sub-image captions to generate the final detailed grounded caption for the entire image. 
This simple but effective strategy of dividing and then merging ensures that the generated image caption contains a sufficient number of objects and details.

\noindent
\textbf{Using SCM.} The computational cost of stage-3 is also huge, as generating sub-image captions and synthesizing the overall image caption both use the 78B model. 
Thus, we also train a small spatial caption merge model (SCM) using the generated 600k images data to accelerate annotation, with details introduced in Sec.~\ref{sec:spatial_caption_merge_model}. 
We use the trained spatial caption merge model to continue annotating 400k images. Through this process, we obtain 1M detailed grounded image captions.




\section{Method}
\label{sec:method}
%
%

\subsection{Detailed Region Caption Model}\label{sec:detailed_region_caption_model}

As shown in Fig.~\ref{fig:object_cap_model}, our DRC is initialized from the Qwen-2.5 VL~\cite{bai2025qwen2} and underwent slight modifications to better understand visual prompt inputs. 
Our region caption model jointly employs feature referencing and ID embedding to make the model more robust in understanding visual prompt inputs.
Unlike previous visual prompts understanding works~\cite{yuan2024osprey, rasheed2024glamm}, which design complex extraction modules to obtain object representations, DRC adopts a more direct and practical approach. Specifically, we crop objects as sub-images from the original image based on the input object masks, which are then fed into Qwen-2.5 VL to provide features of the referenced objects. 
This simple strategy brings two benefits: 1) compared to Osprey~\cite{yuan2024osprey} and GLaMM~\cite{rasheed2024glamm}, referencing objects through sub-images does not introduce additional parameters, thus eliminating the need for complex and carefully designed pre-training, 2) sub-images provide more detailed object information and effectively leverage the powerful multi-image understanding capabilities of Qwen-2.5 VL.

\noindent
\textbf{Adding ID embeddings.}
However, in very challenging scenarios, such as scenes with numerous objects, using only feature references can make the model struggle to identify objects amidst complex scenes and similar object interference accurately. 
For example, Osprey~\cite{yuan2024osprey} is often incorrectly identified as a similar nearby object in dense scenes. To solve this issue, we introduced ID embeddings, which spatially overlay ID embeddings onto corresponding vision features to create a stronger binding relationship between visual prompts and specific image regions. 
Specifically, we randomly assign each object an ID special token $\langle obj \ i \rangle$ and fill its corresponding text embeddings into object masks, passing them through a learnable ID patch embed (implemented via a 2D convolution without patch overlap) and superimposing them onto vision patch embeddings. 
ID embeddings provide a strong spatial binding relationship, and the early fusion of ID embeddings with vision features enables the model to more precisely identify the referenced objects. (See the Appendix~\ref{app_sec:exp_results})
As shown in Tab.~\ref{tab:caption_model}, our proposed region caption model surpasses the Osprey-7 B~\cite {yuan2024osprey} by 8.3 CIDEr and 0.3 METEOR using a 3B model with fewer training data and fewer training stages.

\subsection{Spatial Caption Merging Model}\label{sec:spatial_caption_merge_model}
To accelerate the stage 3 pipeline in ~\ref{fig:pipeline}, we fine-tune a small model to annotate scene-level captions in a single pass based on the given image, object masks, and object captions. 
We adopt the SOTA InternVL3-8B~\cite{zhu2025internvl3} as our merge model, inputting images with visual prompts and object ID tags along with the captions corresponding to each object, and outputting grounded dense image captions. 
Since the task is less challenging, we do not change the model architecture, but directly use the data generated by the stage 3 pipeline for supervised fine-tuning.

\begin{figure}[t]
\hsize=\textwidth
\centering
\includegraphics[width=1.00\textwidth]{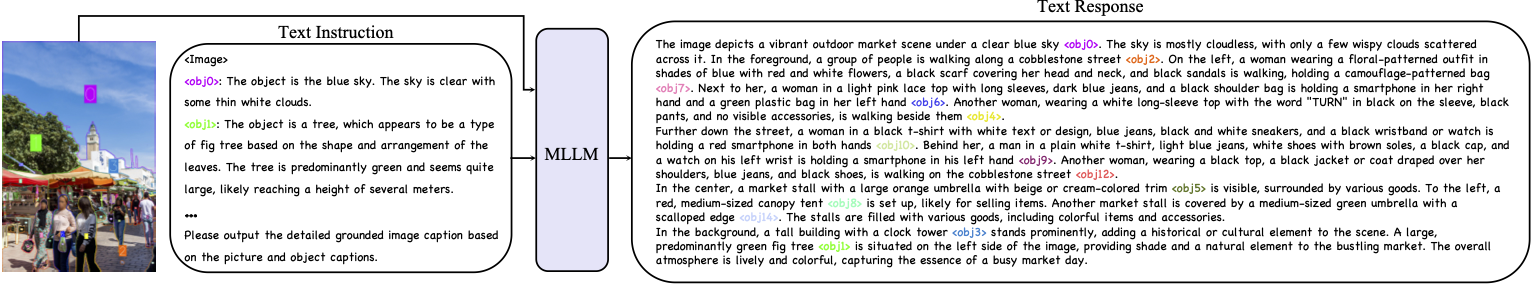}
\caption{\small{\textbf{The proposed Spatial Caption Merging model.}} With multiple inputs, SCM merges multiple detailed object captions into one fluent, dense, grounded caption.}
\label{fig:grounded_image_caption_model}
\end{figure}

\begin{table*}[t!]
    \centering
    \caption{\small{Performance on referring expression segmentation datasets by DenseWorld-1M.}}
    \resizebox{0.85\textwidth}{!}{
    \begin{tabular}{l|c|ccc|ccc|cc}
    \toprule[0.2em]
    \multirow{2}{*}{Method} & \multirow{2}{*}{Size} & \multicolumn{3}{c|}{refCOCO} & \multicolumn{3}{c|}{refCOCO+}  & \multicolumn{2}{c}{refCOCOg} \\
     ~ & ~ & Val & TestA & TestB & Val & TestA & TestB & Val & Test \\
    \midrule
     LISA~\cite{lai2024lisa} & 7B & 74.9 & 79.1 & 72.3 & 65.1 & 70.8 & 58.1 & 67.9 & 70.6 \\
     PixelLM~\cite{ren2024pixellm} & 7B & 73.0 & 76.5 & 68.2 & 66.3 & 71.7 & 58.3 & 69.3 & 70.5 \\
    OMG-LLaVA~\cite{zhang2024omg} & 7B & 78.0 & 80.3 & 74.1 & 69.1  & 73.1 & 63.0 & 72.9  & 72.9 \\
    GLaMM~\cite{rasheed2024glamm} & 7B & 79.5 & 83.2 & 76.9 & 72.6 & 78.7 & 64.6 & 74.2 & 74.9 \\
    \hline
    Sa2VA~\cite{yuan2025sa2va} (Baseline) & 4B & 82.4 & 84.2 & 79.5 & 77.6 & 81.2 & 73.1 & 79.7 & 80.4 \\
    Sa2VA~\cite{yuan2025sa2va} (Baseline) & 8B & 82.7 & 84.6 & 80.0 & 78.0 & 82.0 & 73.8 & 80.2 & 80.3 \\
    Sa2VA (ours) & 4B & 83.2 & 84.9 & 80.5 & 78.8 & 82.3 & 73.7 & 80.0 & 81.0 \\
    Sa2VA (ours) & 8B & \textbf{83.6} & \textbf{85.2} & \textbf{81.6} & \textbf{79.9} & \textbf{83.1} & \textbf{74.9} & \textbf{81.0} & \textbf{81.1} \\
    \bottomrule[0.1em]
    \end{tabular}
    }
    \label{tab:refcoco}
\end{table*}

\begin{table*}[t!]
    \centering
    \caption{\small{Performance on grounded conversation generation dataset by DenseWorld-1M.}}
    \resizebox{0.95\textwidth}{!}{
    \begin{tabular}{l|c|ccccc|ccccc}
    \toprule[0.2em]
    \multirow{2}{*}{Method} & \multirow{2}{*}{Size} & \multicolumn{5}{c|}{Val} & \multicolumn{5}{c}{Test} \\
     ~ & ~ & METEOR & CIDEr & AP$_{50}$ & mIoU & Recall & METEOR & CIDEr & AP$_{50}$ & mIoU & Recall \\
    \midrule
     LISA~\cite{lai2024lisa} & 7B & 13.0 & 33.9 & 25.2 & 62.0 &  36.3 & 12.9 & 32.2 & 24.8 & 61.7 & 35.5 \\
    GLaMM~\cite{rasheed2024glamm} & 7B & 16.2 & 47.2 & 30.8 & 66.3 & 41.8 & 15.8 & 43.5 & 29.2 & 65.6 & 40.8 \\
    MGLMM~\cite{yuan2025instruction} & 7B &  16.4 & 50.1 & 31.7 & 66.3 & 45.2 & - & -& - & - & -\\
    OMG-LLaVA~\cite{zhang2024omg} & 7B & 14.9 &  41.2  & 29.9 & 65.5  & - & 14.5 & 38.5 & 28.6 &  64.7 & - \\
    Sa2VA~\cite{yuan2025sa2va} & 4B & 16.1 & \textbf{53.1} & 31.3 & 68.1 & 43.0 & 15.7 & 50.4 & 31.3 & 67.8 & 44.7 \\
    Sa2VA~\cite{yuan2025sa2va} & 8B & 16.4 & 49.5 & 33.2 & 67.7 & 45.1 & 16.2 & 49.0 & 32.2 & 66.8 & 44.5\\
    \hline 
    Sa2VA (ours) & 4B & 16.3 & 51.5 & \textbf{34.2} & 69.3 & 45.7 & 16.1 & 51.1 & \textbf{34.1} & 68.9 & 46.8 \\
    Sa2VA (ours) & 8B & \textbf{16.9} & 52.5 & 34.0 & \textbf{69.5} & \textbf{47.7} & \textbf{16.8} & \textbf{54.1} & 33.2 & \textbf{69.1} & \textbf{47.8} \\
    \bottomrule[0.1em]
    \end{tabular}
    }
    \label{tab:gcg_exp}
\end{table*}

%% file: sec/4_exp.tex
\section{Experiments}
\label{sec:exp}

\begin{figure}[t]
\hsize=\textwidth
\centering
\includegraphics[width=1.00\textwidth]{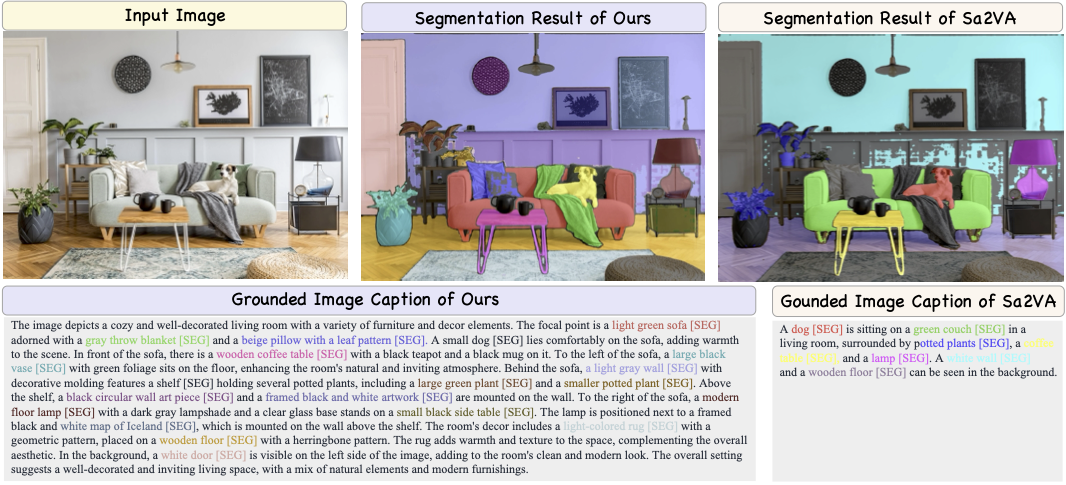}
\caption{\small{\textbf{Visual comparison on dense grounded caption.} With our dataset training, the baseline model can generate more fine-grained and dense grounded captions. Best view it in color.}}
\label{fig:gcg_demos}
\end{figure}

\begin{figure}[t]
\hsize=\textwidth
\centering
\includegraphics[width=1.00\textwidth]{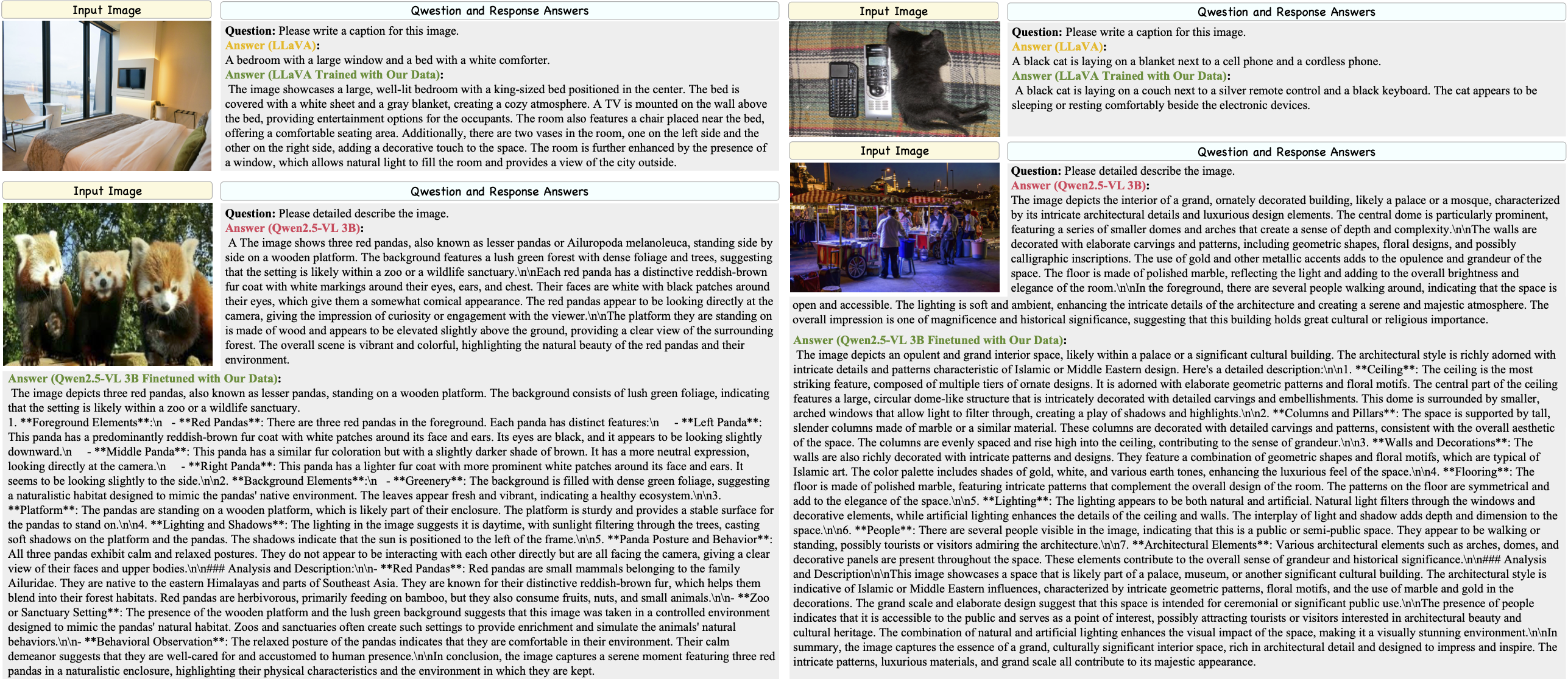}
\caption{\small{\textbf{Visual Comparison on Dense Caption on Image-level MLLMs.} Top: LLaVA. Bottom: Qwen2.5-VL. Both models can generate more detailed, structural captions, compared with the baseline models.}}
\label{fig:caption_demos}
\end{figure}

\noindent
\textbf{Evaluation Benchmarks.} All experiments are carried out to verify the effectiveness of DenseWorld-1M. We mainly verify on standard MLLM benchmarks (MMBench~\cite{mmbench}, MME series~\cite{fu2023mme, zhang2024mme_realworld}, MMStar~\cite{mmstar}, SEEDBench~\cite{li2023seed}, AI2D~\cite{ai2d}, MMVP~\cite{tong2024eyes_mmvp}, MMMU~\cite{mmmu}) and pixel-level MLLM benchmarks (RefCOCO~\cite{yu2016modeling_refcoco_dataset}, RefCOCOg~\cite{refcocog}, RefCOCO+~\cite{refcocog}, Grounded Conversation Generation~\cite{rasheed2024glamm}).

\noindent
\textbf{Baseline Models.} For dataset verification, we adopt two MLLMs and one pixel MLLM. In particular, we first adopt Sa2VA~\cite{yuan2025sa2va} to verify referring segmentation and grounded caption generation ability, since these tasks contain both mask and text prediction. 
Then, we use Qwen-2.5 VL as post-training models, and we adopt LLaVA-1.5 for training-from-scratch verification. For LLaVA, we built a strong baseline, which uses Qwen-2.5 3B as the LLM while keeping other settings, including training strategies and data, consistent with LLaVA 1.5.

\noindent
\textbf{Implementation details.} Due to limited pages, we refer readers to the Appendix~\ref{app_sec:exp_results} for reference.

\subsection{Main Results}
\label{sec:main_results}

\noindent
\textbf{Main Results For Pixel-level MLLMs.} In Tab.~\ref{tab:refcoco}, we first verify the effectiveness on referring segmentation tasks. With our dense grounded captions fine-tuning, we find about 0.5\% -1 \% improvements over a strong baseline, Sa2VA~\cite{yuan2025sa2va}.
In Tab.~\ref{tab:gcg_exp}, for the grounded caption generation task, we find improvements on multiple metrics, including caption, grounded caption recall, and masks. 
In particular, we find more significant improvements on grounded captions (AP50, Recall), indicating the effectiveness of our data, despite huge domain gaps.
The METEOR and CIDEr are not.
We argue that these metrics are more sensitive to the format and length of outputs, while the GCG dataset has a short text description.

\noindent
\textbf{Main Results For Image-level MLLMs.} We verify the effectiveness of our DenseWorld-1M dataset on Image-level MLLMs. We first verify it with the current STOA Qwen2.5-VL, with post-training. As shown in Tab.~\ref{tab:exp_qwen_2_5_vl}, we can still find consistent improvements over eight different datasets.
This indicates that our dataset will benefit the MLLM community for building stronger baselines.
In addition, we also perform LLaVA-1.5 experiment to verify the pre-training effect of DesneWorld-1M, where we adopt Qwen-2.5 LLM as the language backbone.
As shown in Tab.~\ref{tab:llava_baseline_exp}, we can still find consistent improvements on many real-world VQA benchmarks. 
Please find more results and details in the Appendix~\ref{app_sec:exp_results}.

\noindent
\textbf{Effectiveness of DRC.} In Tab.~\ref{tab:caption_model}, we verify the effectiveness of DRC under the previous setting. 
In particular, we do not use our DenseWorld-1M and use the Osprey datasets for fair comparison. 
Without any bells and whistles, DRC achieves SOTA results, compared with recent MLLMs.
Due to page limitations, we refer the readers to Appendix~\ref{app_sec:exp_results} for the specific ablation on DRC.

\noindent
\textbf{Effectiveness of SPM.} Since there are no existing benchmarks to evaluate the caption merging ability. 
Thus, we carry out a user study by randomly selecting 100 examples, where our stage-3 pipeline and SPM infer these examples. 
We have asked over 20 people to choose which one is better.
As a result, we find that both results are very close.
We refer the readers to Appendix~\ref{app_sec:exp_results} for the detailed results.

\noindent
\textbf{Efficiency of DRC and SCM.} We calculate the labeling efficiency of the pipelines, DRC, and SCM on 1000 SAM images, with all models using LMDeploy~\cite{2023lmdeploy} for inference acceleration. In the stage-2 labeling process, on an 80GB A100 GPU, the labeling pipeline requires an average of 3.2 minutes to annotate all objects in a SAM-level image, while DRC only needs 1.1 minutes. In the stage-3 labeling process, the labeling pipeline requires an average of 2.6 minutes to generate dense grounded captions for a SAM-level image, while SCM only needs 31 seconds. All experiments are conducted on one A100 GPU.

\subsection{Visual Comparison}
\label{exp:visual_comparison}

\noindent
\textbf{Visual comparison on Dense Grounded Caption.} In Fig.~\ref{fig:gcg_demos}, we present detailed grounded caption generation. With our DenseWorld-1M fine-tuning, Sa2VA can generate dense masks with more detailed words to describe the scene compared with the original baseline. 

\noindent
\textbf{Visual comparison on Dense Caption.} We present several examples in image-level MLLM in Fig.~\ref{fig:caption_demos}. 
For both LLaVA and Qwen2.5-VL, after training with DenseWorld-1M, both models can generate detailed captions to describe object contents, object relations, and locations.
Please find more demos in Appendix~\ref{app_sec:more_visual_examples}.

\begin{table*}[t!]
    \centering
    \caption{\small{Performance on MLLM benchmarks using Qwen2.5-VL as strong baseline.}}\vspace{-2mm}
    \resizebox{0.95\textwidth}{!}{
    \begin{tabular}{c|c|cccccccc}
    \toprule[0.2em]
    Method & LLM-Model-Size & AI2D & MMBench 1.1 & MMStar & MME &  SEEDBench & MME-RealWorld & MMVP & MMMU \\
    \midrule
     Qwen2.5-VL & 3B & 78.4 & 76.5 & 54.9 & \textbf{2200} & 73.9 & 55.2 & 67.3 & 47.1 \\
    \hline 
     + DenseWorld-1M & 3B & \textbf{81.7} & \textbf{78.0} & \textbf{58.1} & {2175} & \textbf{74.8} & \textbf{58.7} & \textbf{69.3} & \textbf{47.4} \\
    \bottomrule[0.1em]
    \end{tabular}
    }
    \label{tab:exp_qwen_2_5_vl}\vspace{-2mm}
\end{table*}

\begin{table*}[t!]
    \centering
    \caption{\small{Performance on MLLM benchmarks using LLaVA as baseline.}}\vspace{-2mm}
    \resizebox{0.92\textwidth}{!}{
    \begin{tabular}{c|c|cccccccc}
    \toprule[0.2em]
    Method & Size & MMBench & MME & MMStar & SEEDBench & AI2D & MMVP & HallusionBench & MMMU \\
    \midrule
     LLaVA 1.5 & 3B & 67.0 & 1686 & 41.7 & 69.4 & 63.4 & 62.7 & 47.6 &  36.9 \\
    \hline 
     + DenseWorld-1M & 3B & \textbf{73.6} & \textbf{1801} & \textbf{45.3} & \textbf{73.9} &\textbf{65.0} & \textbf{65.3} & \textbf{50.4} & \textbf{44.8} \\
    \bottomrule[0.1em]
    \end{tabular}
    }
    \label{tab:llava_baseline_exp}\vspace{-2mm}
\end{table*}

\begin{table*}[t!]
    \centering
    \caption{\small{Performance of region captioning on the RefCOCOg dataset. To demonstrate the architectural superiority of our region captioning model, we use only Osprey-724K~\cite{yuan2024osprey} for supervised fine-tuning without any meticulously designed pre-training stages.}}\vspace{-2mm}
    \resizebox{0.90\textwidth}{!}{
    \begin{tabular}{c|c|cccccccc}
    \toprule[0.2em]
    Results & DRC (ours) & Osprey & GLaMM & RegionGPT & Groma & ViP-LLaVA & Kosmos-2 & GRIT \\
    \hline
    CIDEr & \textbf{116.6} & 108.3 & 105.0 & 109.9 & 107.3 & 105.9 & 62.3  & 71.6  \\
    METEOR & \textbf{16.9} & 16.6 & 16.2 & 16.9 & 16.8 & 16.6 & 14.1 & 15.2 \\
    \bottomrule[0.1em]
    \end{tabular}
    }
    \label{tab:caption_model}\vspace{-2mm}
\end{table*}

%% file: sec/5_conclusion.tex
\vspace{-2mm}\section{Conclusion}\vspace{-2mm}
\label{sec:conclusion}

In this work, we introduce DenseWorld-1M, the first large-scale, high-detail dense grounded caption dataset for real-world scenes. 
Our three-stage labeling pipeline—open-world perception, detailed object caption generation, and dense caption merging—first extracts entity-level masks/labels, then generates object-level descriptions guided by these annotations, and finally fuses captions and masks into spatially relational dense captions. 
To optimize labeling efficiency and caption quality, we develop two MLLMs: the Detailed Region Caption (DRC) model and the Spatial Caption Merging (SCM) model. 
Extensive experiments across vision-language understanding, visual grounding, and region caption generation tasks validate the effectiveness of DenseWorld-1M and our labeling models, with results also demonstrating DRC/SCM utility. 
Our work aims to inspire the MLLM community to pursue more fine-grained real-world understanding.

\noindent
\textbf{Limitations and Future Works.} Our work still has room to improve. 
From the data source perspective, we can include more diverse data formats, including video data, synthetic data, and text-to-image datasets. 
From a data scale perspective, we will further enlarge our data engine to the 10M scale. 
In addition, our dataset can also be used for O3-like dataset building, such as step-by-step visual reasoning.

%% file: sec/X_suppl.tex
\clearpage
\beginappendix

\noindent
\textbf{Overview.} This appendix contains four parts: We provide statistics on the DenseWorld dataset attributes in Sec. ~\ref{app_sec:dataset_details}, along with visualization results of more samples. We present more validation experiments on larger models in Sec.~\ref{app_sec:exp_results}. In Sec.~\ref{app_sec:more_visual_examples}, we provide more prediction results of models trained on our data. In Sec.~\ref{app_sec:more_implementation_details}, we introduce additional implementation details.

\section{Appendix For More Dataset Details}
\label{app_sec:dataset_details}

\begin{figure}[t]
    \centering
    \includegraphics[width=1.0\linewidth]{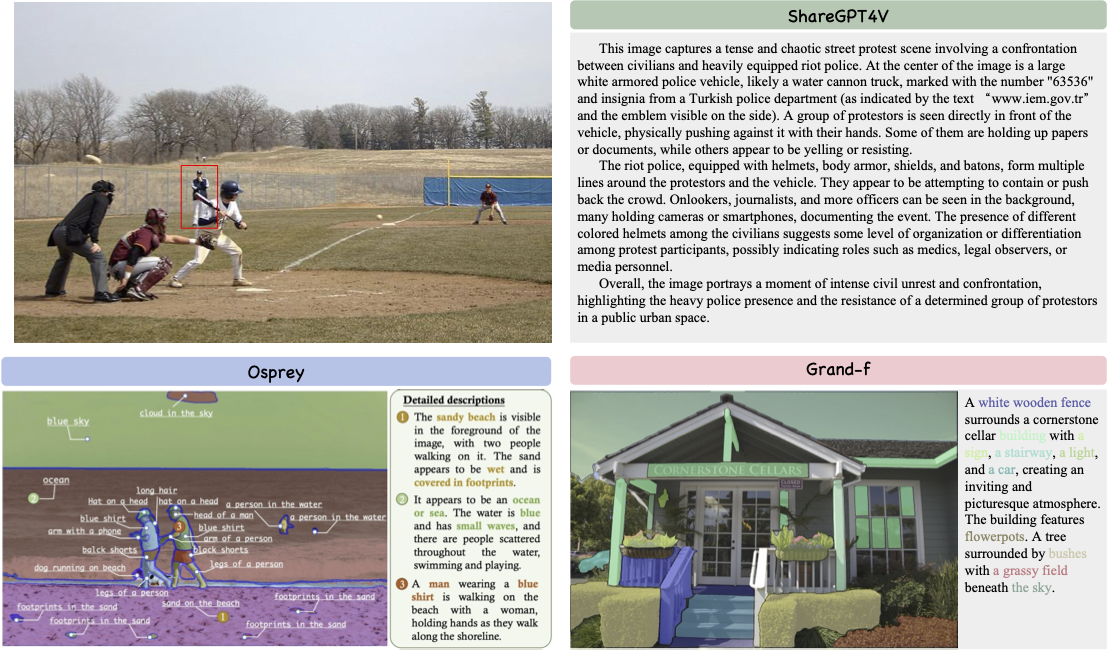}
    \caption{\textbf{Visualization examples from other datasets}. ShareGPT4V tends to miss objects in the scene, such as the person highlighted in the red box, while the captions from both Osprey and GranD-f datasets are very brief, containing extremely limited information.}\label{fig:other_dataset_sample}
\end{figure}

\noindent\textbf{Statistical information.}
We analyze the properties of captions in DenseWorld, including the number of characters, words, and sentences, with results shown in Tab.~\ref{tab:statistical_information}. Compared to ShareGPT4V~\cite{chen2024sharegpt4v} and DenseFusion~\cite{li2024densefusion}, our scene-level captions contain nearly twice as many sentences and more than twice as many words. Additionally, our DenseWorld includes 23.1M object captions and 23.6M object mask annotations. Our DenseWorld surpasses previous datasets in terms of annotation granularity, number of samples, caption detail, and degree of alignment between text and masks, as shown in Fig.~\ref{fig:other_dataset_sample}.

\begin{table*}[h]
    \centering
    \caption{\small{\textbf{Statistical information on Denseworld-1M.} "Char." represents the average number of characters in captions, "Word" represents the average number of words in captions, and "Sen." represents the average number of sentences in captions. "-" indicates that the paper did not report the corresponding statistical results, "0" indicates that the dataset does not contain this type of data.}}
    \resizebox{0.9\textwidth}{!}{
    \begin{tabular}{c|cccc|cccc|c}
    \toprule[0.2em]
    \multirow{2}{*}{Method} & \multicolumn{4}{c|}{Scene-Level} & \multicolumn{4}{c|}{Object-Level} & \multicolumn{1}{c}{Pixel-Level} \\
    ~ & Samples & Char. & Word & Sen. & Samples & Char. & Word & Sen. & Samples\\
    \midrule
    ShareGPT4V~\cite{chen2024sharegpt4v} & 0.1M & 942 & - & - & 0 & 0 & 0 & 0 & 0\\
    DenseFusion~\cite{li2024densefusion} & 1M & 1253 & 206 & 11.2 & 0 & 0 & 0 & 0 & 0 \\
    \hline
     DenseWorld & 1M & 2813.1 & 458.4 & 20.5 & 23.1M & 673.4 & 111.4 & 4.5 & 23.6M \\
    \bottomrule[0.1em]
    \end{tabular}
    }
    \label{tab:statistical_information}
\end{table*}

\begin{figure}[t]
    \centering
    \includegraphics[width=0.9\linewidth]{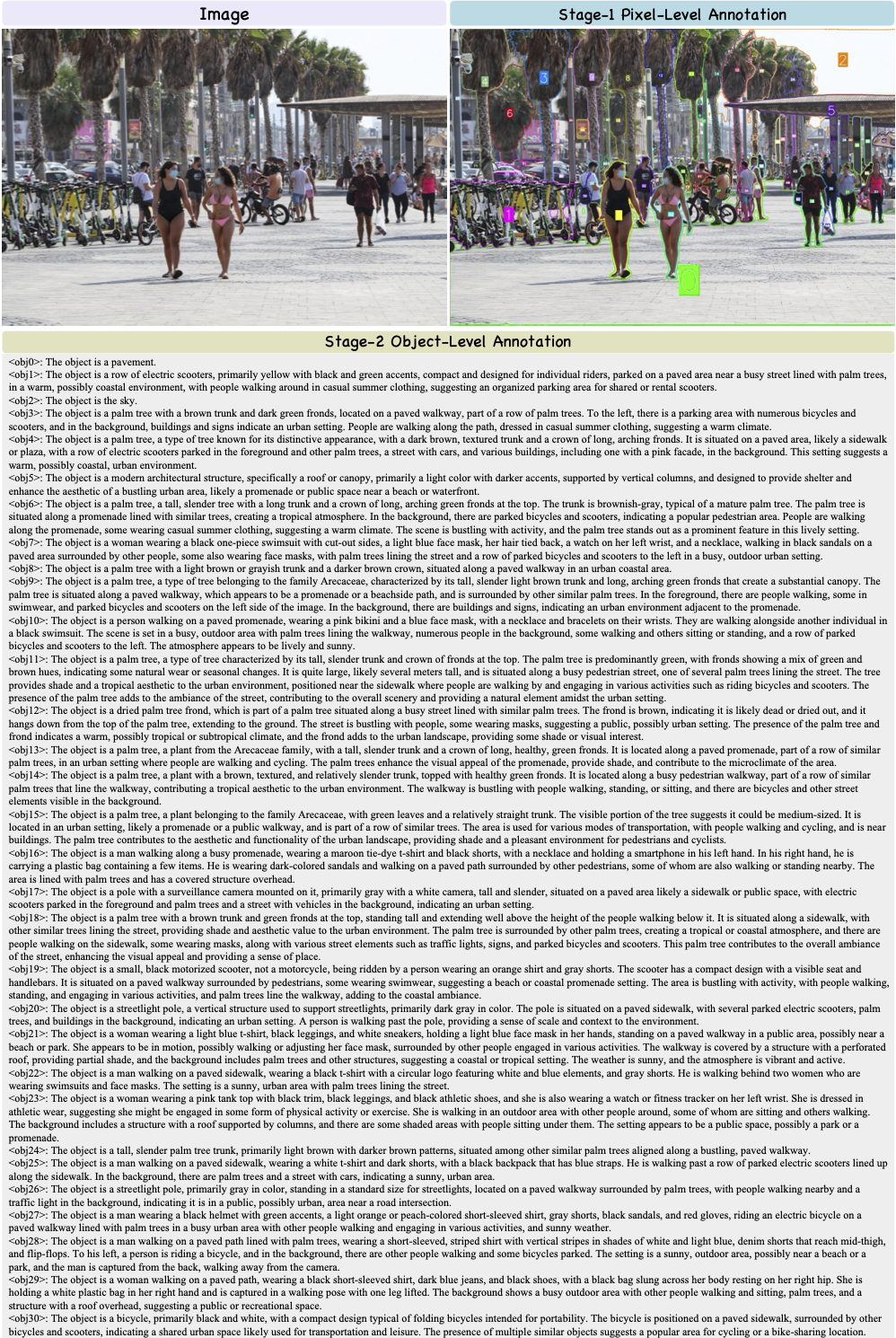}
    \caption{\textbf{More visualization results of DenseWorld dataset. Part 1 of sample 1.}} 
    \label{fig:more_data_samples_1_1}
\end{figure}

\begin{figure}[t]
    \centering
    \includegraphics[width=0.90\linewidth]{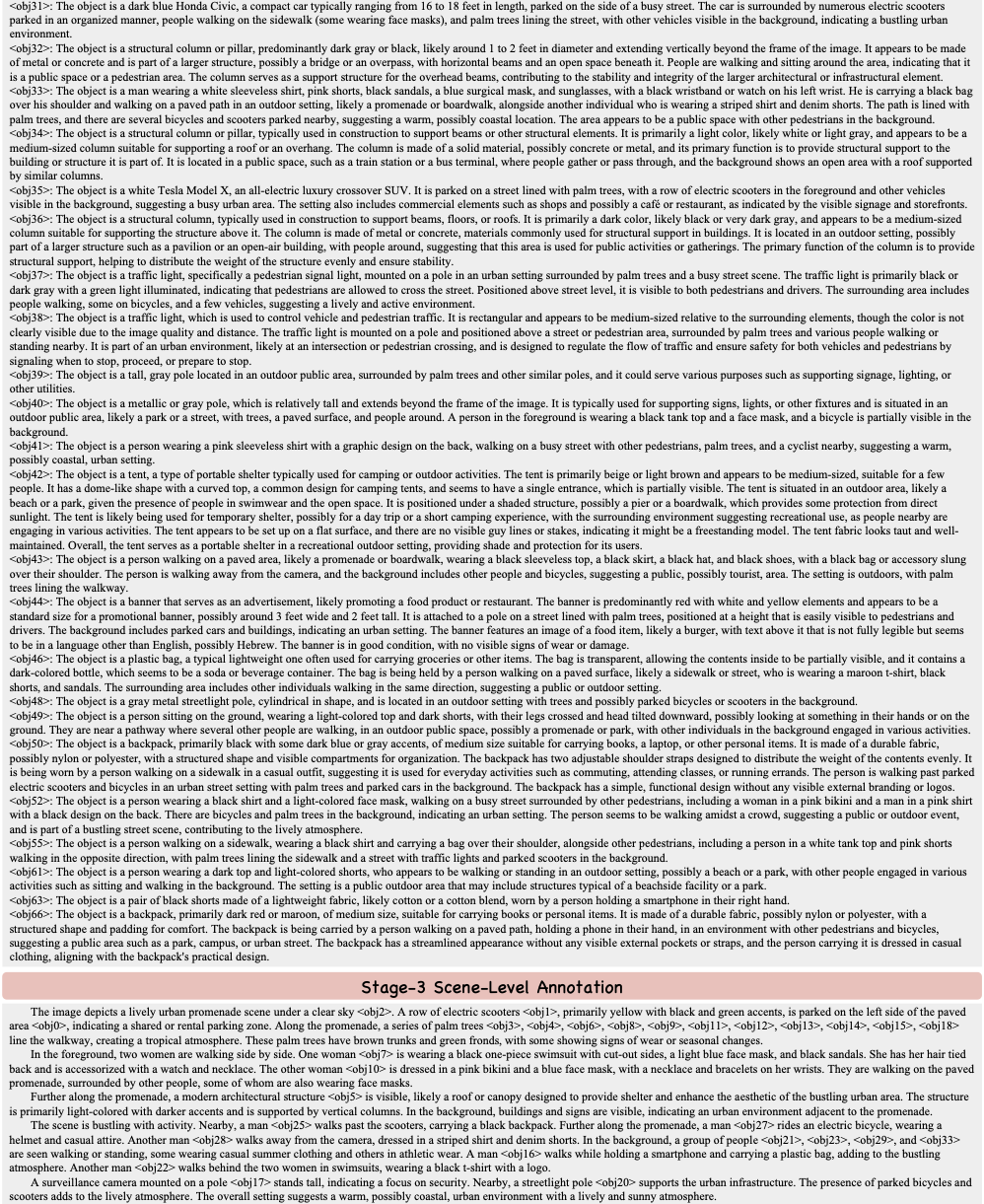}
    \caption{\textbf{More visualization results of DenseWorld dataset. Part 2 of sample 1.}} 
    \label{fig:more_data_samples_1_2}
\end{figure}

\begin{figure}[t]
    \centering
    \includegraphics[width=0.90\linewidth]{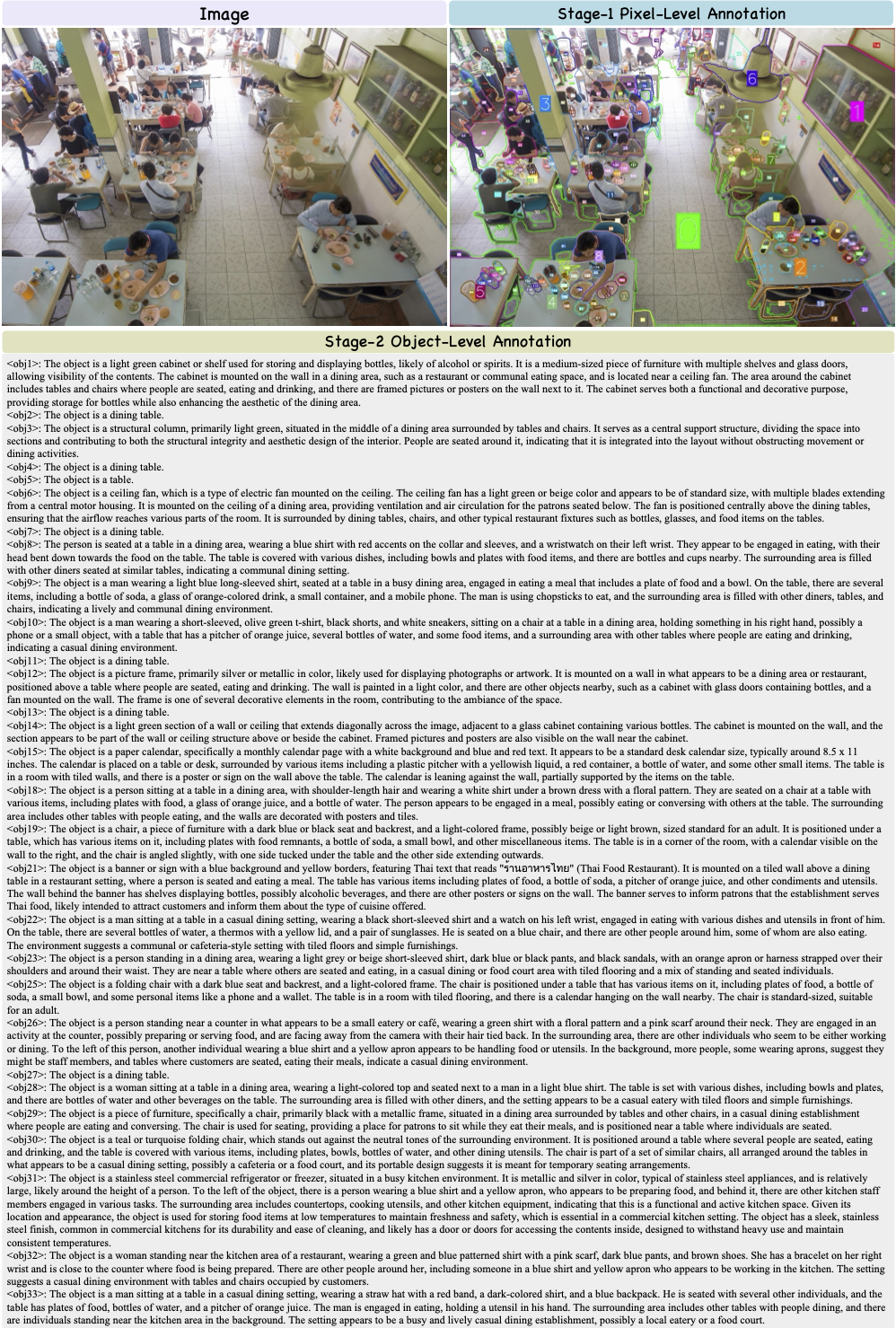}
    \caption{\textbf{More visualization results of DenseWorld dataset. Part 1 of sample 2.}} 
    \label{fig:more_data_samples_2_1}
\end{figure}

\begin{figure}[t]
    \centering
    \includegraphics[width=0.82\linewidth]{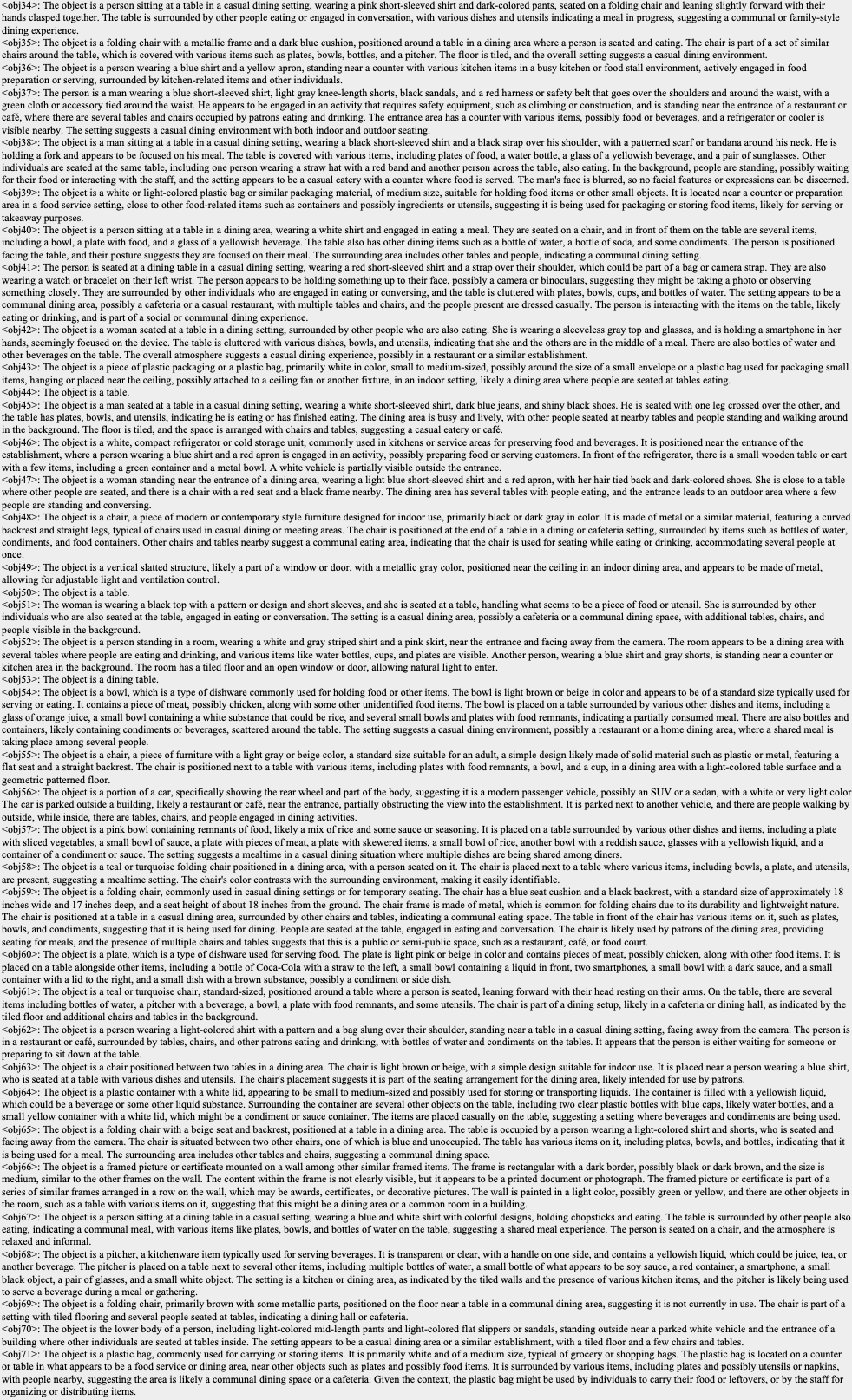}
    \caption{\textbf{More visualization results of DenseWorld dataset. Part 2 of sample 2.}} 
    \label{fig:more_data_samples_2_2}
\end{figure}

\begin{figure}[t]
    \centering
    \includegraphics[width=0.9\linewidth]{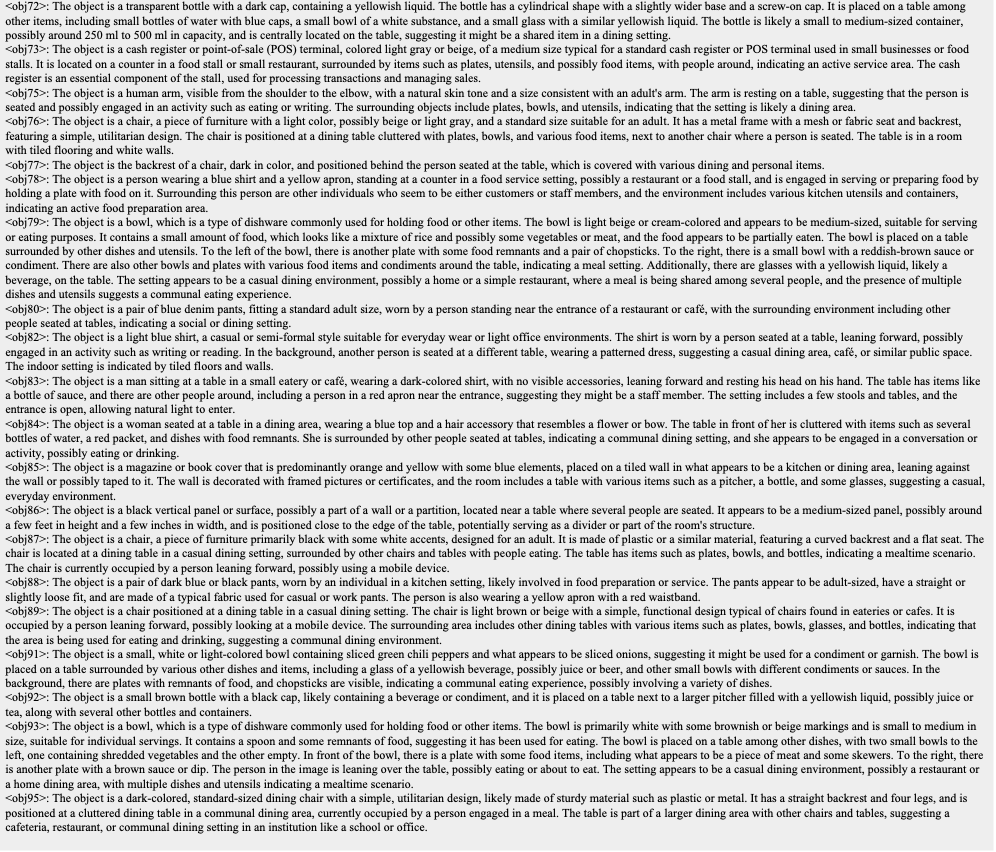}\vspace{-3mm}
    \includegraphics[width=0.9\linewidth]{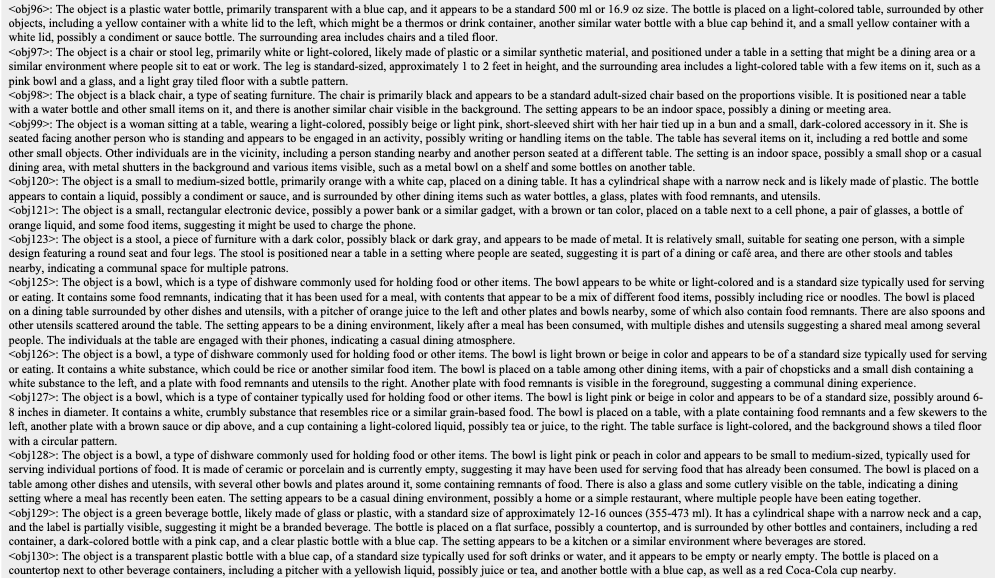}
    \caption{\textbf{More visualization results of DenseWorld dataset. Part 3 of sample 2.}} 
    \label{fig:more_data_samples_2_3}
\end{figure}

\begin{figure}[t]
    \centering
    \includegraphics[width=0.9\linewidth]{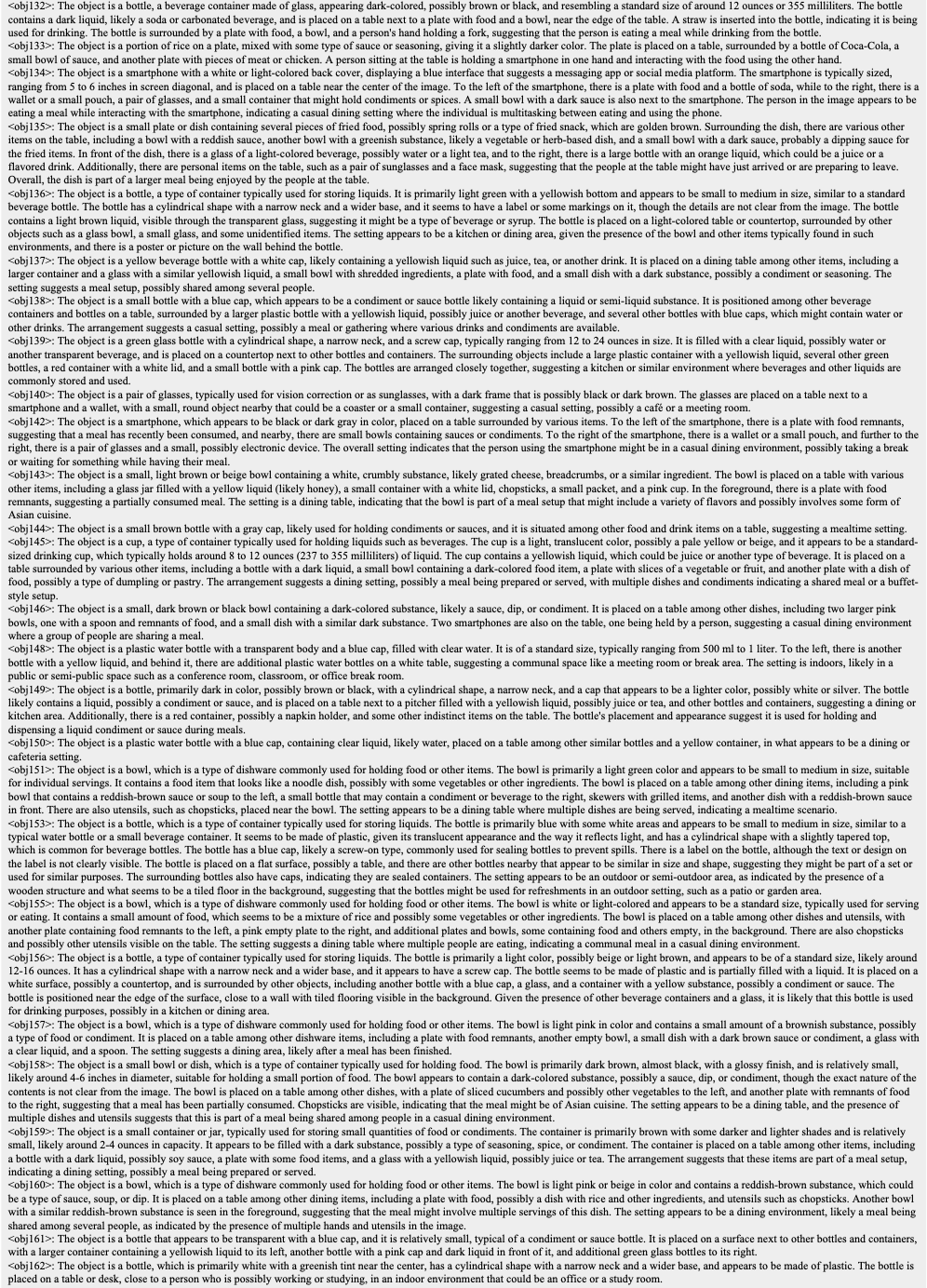}
    \caption{\textbf{More visualization results of DenseWorld dataset. Part 4 of sample 2.}} 
    \label{fig:more_data_samples_2_4}
\end{figure}

\begin{figure}[t]
    \centering
    \includegraphics[width=0.9\linewidth]{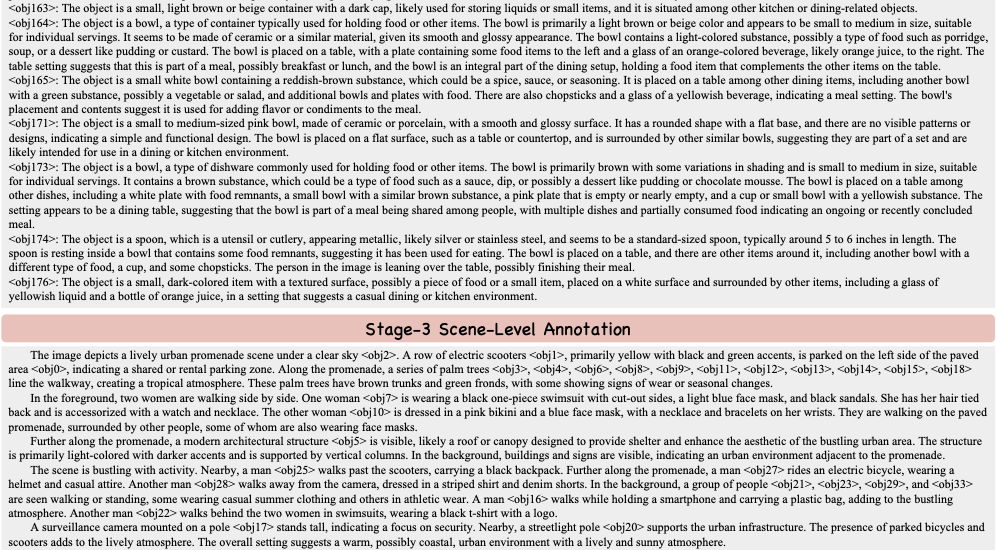}
    \caption{\textbf{More visualization results of DenseWorld dataset. Part 5 of sample 2.}} 
    \label{fig:more_data_samples_2_5}
\end{figure}

\noindent
\textbf{More visualization results.}
We provide a comprehensive display of more samples from the DenseWorld dataset, as shown in Fig.~\ref{fig:more_data_samples_1_1}, \ref{fig:more_data_samples_1_2}, \ref{fig:more_data_samples_2_1}, \ref{fig:more_data_samples_2_2}, \ref{fig:more_data_samples_2_3}, \ref{fig:more_data_samples_2_4}, and \ref{fig:more_data_samples_2_5}. DenseWorld contains abundant dense scenes and provides detailed and accurate pixel-level, object-level, and scene-level annotations.

\noindent
\textbf{Previous dataset examples.} In Fig.~\ref{fig:other_dataset_sample}, we show several examples of previous caption and pixel caption datasets. These features motivate us to build the DenseWorld-1M dataset.

\begin{figure}[t]
    \centering
    \includegraphics[width=1.0\linewidth]{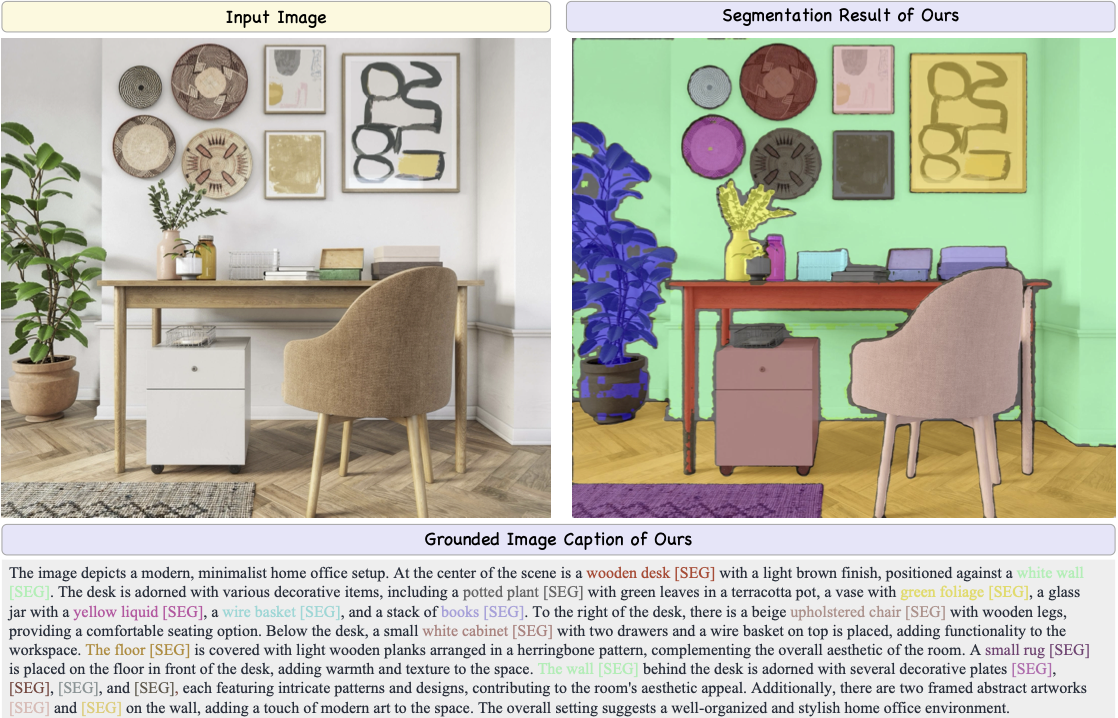}
    \includegraphics[width=1.0\linewidth]{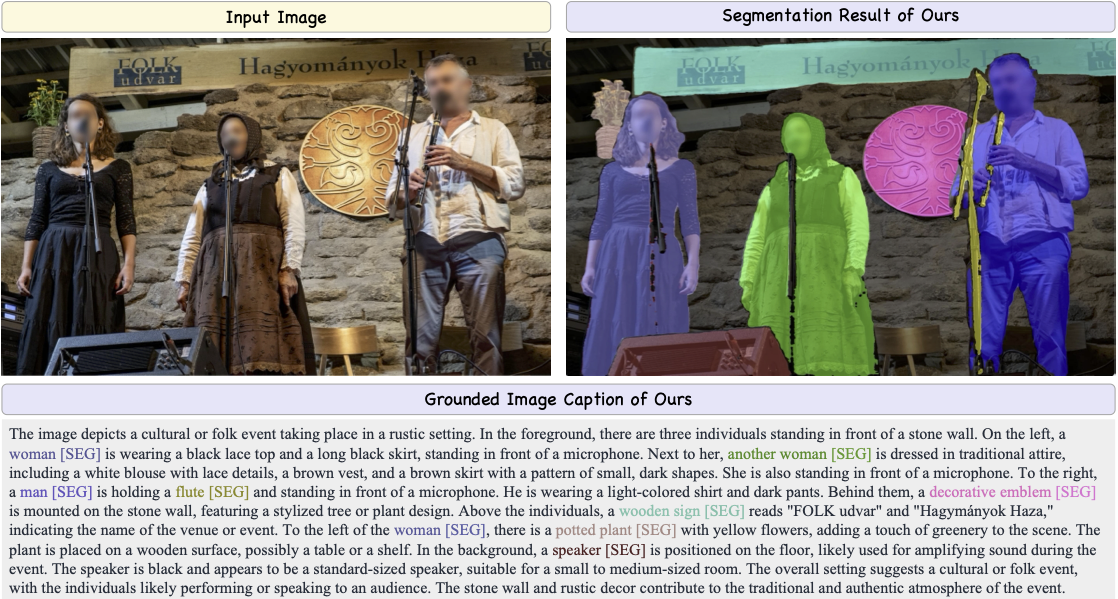}
    \caption{\textbf{More prediction results of the grounded conversation generation task.}} 
    \label{fig:more_gcg_demos}
\end{figure}

\begin{figure}[t]
    \centering
    \vspace{-5mm}\includegraphics[width=0.82\linewidth]{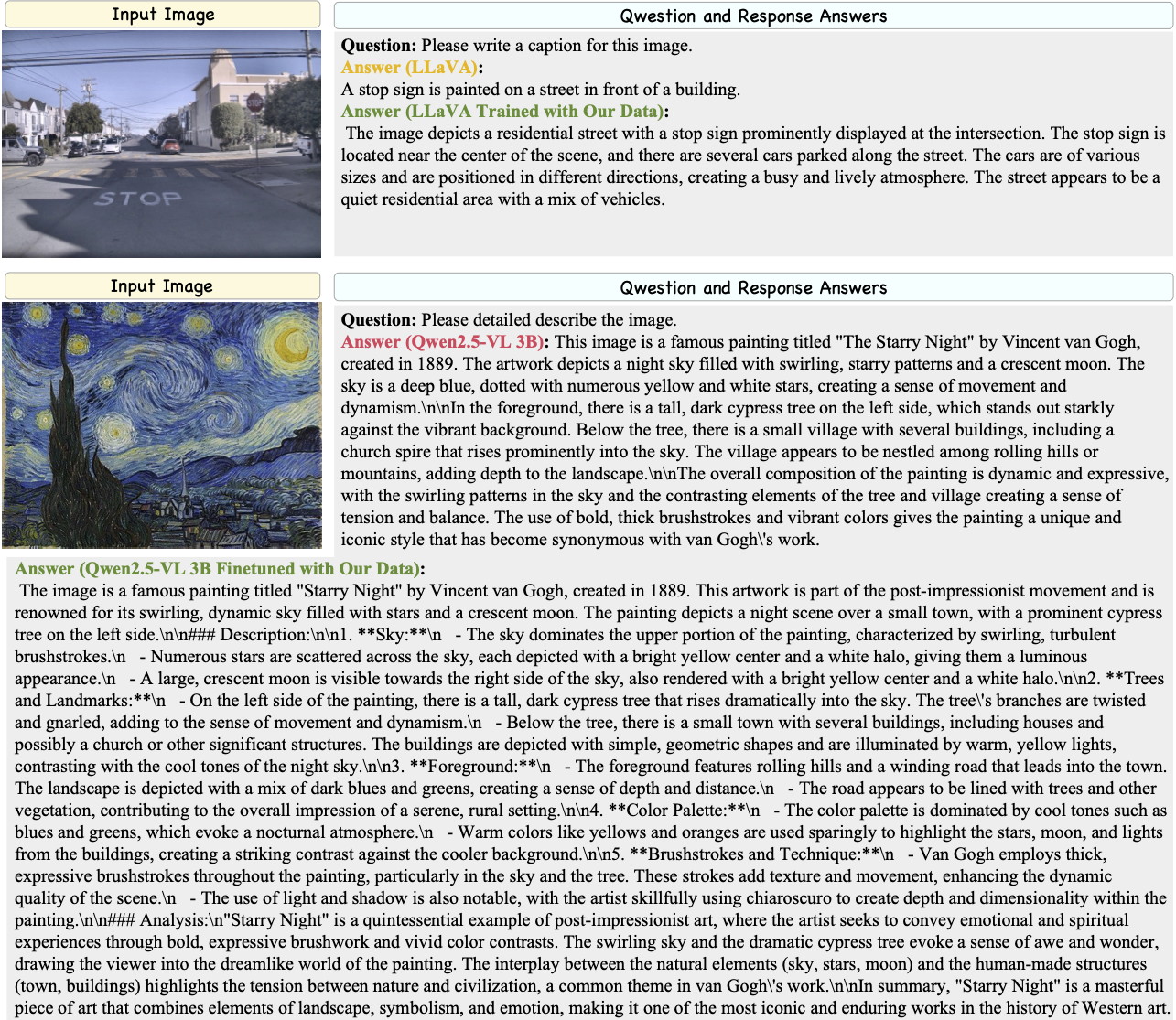}
    \includegraphics[width=0.82\linewidth]{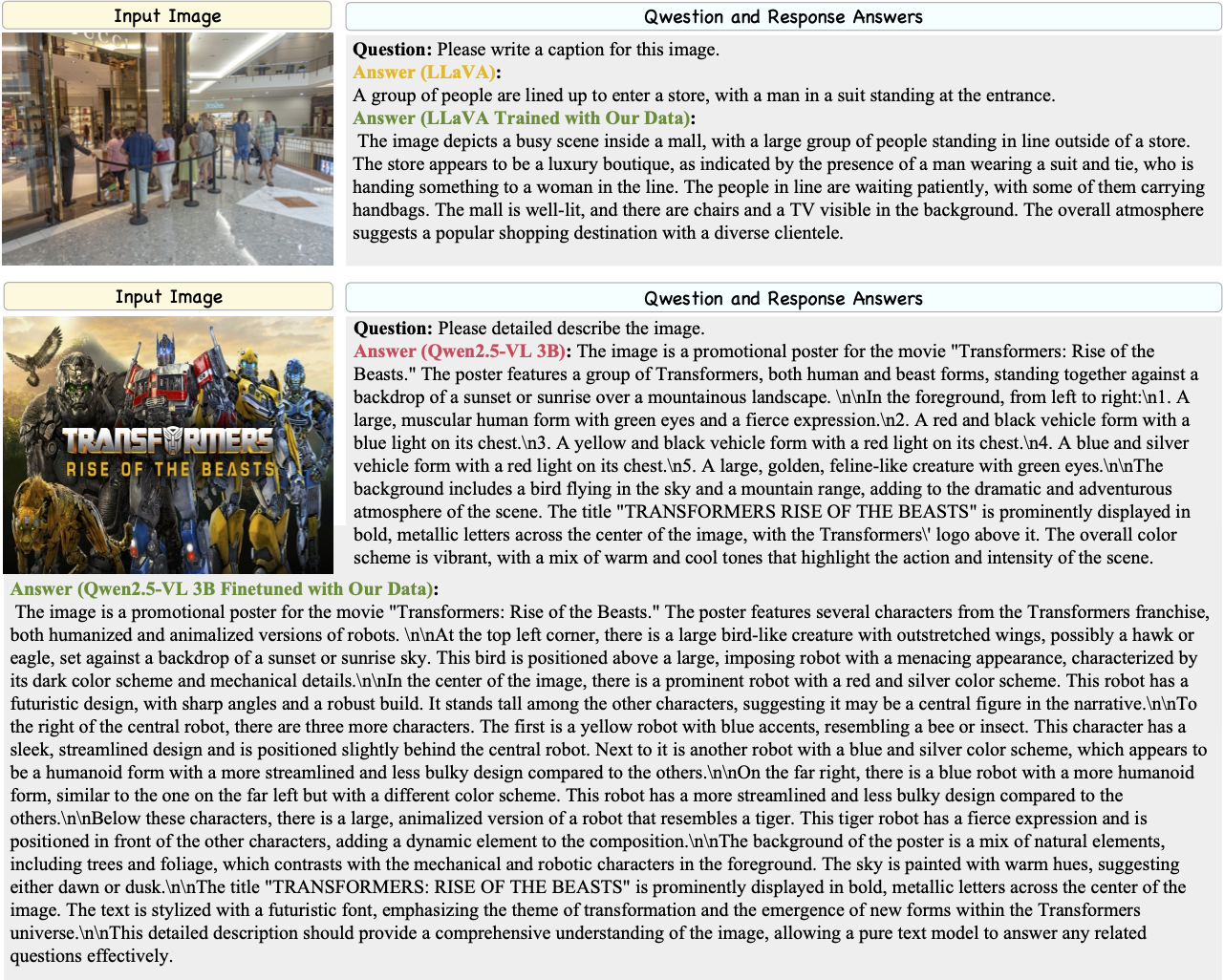}
    \caption{\textbf{More prediction results of the image caption task.}} 
    \label{fig:more_caption_demos}
\end{figure}

\section{Appendix For More Experiments}
\label{app_sec:exp_results}

\noindent\textbf{LLaVA.} We validate our DenseWorld dataset on LLaVA~\cite{liu2023visual, liu2024improved} with larger LLMs, with results shown in Tab.~\ref{tab:llava_baseline_exp_7b}. When trained with DenseWorld data, the model demonstrates significant improvements on nearly all benchmarks, including general benchmarks, hallucination benchmarks, and reasoning benchmarks. 

\noindent\textbf{Sa2VA.} We validate the DenseWorld dataset on Sa2VA~\cite{yuan2025sa2va} with larger LLMs, with results shown in Tab.~\ref{tab:gcg_exp_8b}. When using the DenseWorld dataset, Sa2VA achieves 0.9, 1.9, and 0.8 cIoU improvements on the validation sets of RefCOCO, RefCOCO+, and RefCOCOg, respectively. Additionally, Sa2VA demonstrates improvements of 0.6 METEOR, 5.1 CIDEr, 1.0 AP50, 2.3 mIoU, and 3.3 Recall on the grounded conversation generation task.

\section{Appendix For More Visual Examples.}
\label{app_sec:more_visual_examples}

We provide additional visual examples of the model trained with our data, as shown in Fig.~\ref{fig:more_gcg_demos} and~\ref{fig:more_caption_demos}. Our DenseWorld data significantly improves the baseline's performance on both grounded conversation generation and dense image caption tasks.


\begin{table*}[t!]
    \centering
    \caption{\small{\textbf{Performance on MLLM benchmarks using LLaVA as baseline.}}}\vspace{-2mm}
    \resizebox{0.92\textwidth}{!}{
    \begin{tabular}{c|c|cccccccc}
    \toprule[0.2em]
    Method & Size & MMBench & MME & MMStar & SEEDBench & AI2D & MMVP & HallusionBench & MMMU \\
    \midrule
     LLaVA 1.5 & 7B & 75.1 & \textbf{1943} & 44.7 & 71.4 & 67.8 & 64.7 & 48.8 & 42.9 \\
    \hline 
     + DenseWorld-1M & 7B & \textbf{77.4} & 1909 & \textbf{48.8} & \textbf{75.4} & \textbf{72.1} & \textbf{72.7} & \textbf{51.2} & \textbf{47.4}  \\
    \bottomrule[0.1em]
    \end{tabular}
    }
\label{tab:llava_baseline_exp_7b}\vspace{-2mm}
\end{table*}

\begin{table*}[t!]
    \centering
    \caption{\small{\textbf{Performance on referring expression segmentation datasets by DenseWorld-1M.}}}
    \resizebox{0.8\textwidth}{!}{
    \begin{tabular}{l|c|ccc|ccc|cc}
    \toprule[0.2em]
    \multirow{2}{*}{Method} & \multirow{2}{*}{Size} & \multicolumn{3}{c|}{refCOCO} & \multicolumn{3}{c|}{refCOCO+}  & \multicolumn{2}{c}{refCOCOg} \\
     ~ & ~ & Val & TestA & TestB & Val & TestA & TestB & Val & Test \\
    \midrule
    Sa2VA~\cite{yuan2025sa2va} (Baseline) & 8B & 82.7 & 84.6 & 80.0 & 78.0 & 82.0 & 73.8 & 80.2 & 80.3 \\
    Sa2VA (ours) & 8B & \textbf{83.6} & \textbf{85.2} & \textbf{81.6} & \textbf{79.9} & \textbf{83.1} & \textbf{74.9} & \textbf{81.0} & \textbf{81.1} \\
    \bottomrule[0.1em]
    \end{tabular}
    }
    \label{tab:refcoco_8b}
\end{table*}

\begin{table*}[t!]
    \centering
    \caption{\small{\textbf{Performance on grounded conversation generation dataset by DenseWorld-1M.}}}
    \resizebox{0.95\textwidth}{!}{
    \begin{tabular}{l|c|ccccc|ccccc}
    \toprule[0.2em]
    \multirow{2}{*}{Method} & \multirow{2}{*}{Size} & \multicolumn{5}{c|}{Val} & \multicolumn{5}{c}{Test} \\
     ~ & ~ & METEOR & CIDEr & AP$_{50}$ & mIoU & Recall & METEOR & CIDEr & AP$_{50}$ & mIoU & Recall \\
    \midrule
    Sa2VA~\cite{yuan2025sa2va} & 8B & 16.4 & 49.5 & 33.2 & 67.7 & 45.1 & 16.2 & 49.0 & 32.2 & 66.8 & 44.5\\
    \hline 
    Sa2VA (ours) & 8B & \textbf{16.9} & \textbf{52.5} & \textbf{34.0} & \textbf{69.5} & \textbf{47.7} & \textbf{16.8} & \textbf{54.1} & \textbf{33.2} & \textbf{69.1} & \textbf{47.8} \\
    \bottomrule[0.1em]
    \end{tabular}
    }
    \label{tab:gcg_exp_8b}
\end{table*}

\section{Appendix For More Implementation details.}
\label{app_sec:more_implementation_details}

\noindent\textbf{Sa2VA.} We use the segmentation-related data from DenseWorld to further train Sa2VA~\cite{yuan2025sa2va} to validate the effectiveness of our data. We utilize the grounded caption data generated from stage-3 as well as the object caption data from stage-2. For the object caption data from stage-2, we reorganize it into referring expression segmentation format: \textit{"Please segment the object. The object is...".} The training settings remain completely consistent with Sa2VA's SFT stage, including learning rate, batch size, warm-up ratio, gradient norm, and other hyperparameters. Since there exist gaps between our data and public benchmarks, such as the use of short expressions in RefCOCO and similarly concise captions in GCG, we further fine-tune the Sa2VA model trained on our data using the training sets of downstream tasks to bridge this gap and fairly evaluate the effectiveness of our data. For task-specific finetuning, we employ exactly the same settings as the original Sa2VA's finetuning protocol for specific tasks.

\noindent\textbf{LLaVA.} To validate our dense caption data produced in stage-3, we train a LLaVA~\cite{liu2024improved, liu2023visual} model from scratch, similar to the approaches used by ShareGPT4V~\cite{chen2024sharegpt4v} and DenseFusion~\cite{li2024densefusion}. To fully demonstrate the effectiveness of our data, we first construct a strong baseline LLaVA by incorporating widely validated community practices. Specifically, we adopt the strong LLM Qwen2.5~\cite{yang2025qwen3}. Second, to avoid over-compressing image resolution and significant information loss, we employ the image partitioning strategy used by the InternVL series~\cite{zhu2025internvl3, chen2024internvl2_5, chen2024expanding}, which preserves high-resolution image information as much as possible through a series of small sub-images and a downscaled thumbnail. For training, we adopt the same two-stage training strategy as LLaVA, including pretraining and SFT. In the pretraining stage, only the MLP remains unfrozen. In the SFT stage, both the vision encoder and LLM are fine-tuned using LoRA~\cite{hu2022lora}, with all MLP parameters being trained as well. Using exactly the same training data as LLaVA 1.5~\cite{liu2024improved}, our strong baseline LLaVA achieves performance that significantly surpasses LLaVA 1.5~\cite{liu2024improved}.

We incorporate our dense caption data into the pretraining stage to validate its effectiveness. Specifically, we first use LLaVA pretraining data to train only the MLP, then use our dense caption data to fine-tune the vision encoder and LLM with LoRA. For the SFT stage, we use exactly the same settings and data as our baseline.

\textbf{Qwen2.5 VL.} Since Qwen 2.5 VL~\cite{bai2025qwen2} does not open-source its training data, we can only validate the effectiveness of DenseWorld data through further SFT. We mix our dense caption data with MAmmoTH-VL-Instruct-12M~\cite{guo2024mammoth} data at a 1:2 ratio, totaling 3M data, to serve as SFT data. We conduct further fine-tuning on Qwen 2.5 VL using the 3M SFT data, following the default training parameter settings from the official repository. After training, we conduct testing using VLMEvalKit without using any APIs to assist with answer processing.
